\documentclass{article}
\usepackage{graphicx} % Required for inserting images

\usepackage[utf8]{inputenc}
\usepackage[T1]{fontenc}

\usepackage{amsmath,amssymb,amsfonts,graphicx}
\usepackage{multirow}
\usepackage{setspace}
\usepackage{tocloft}
\usepackage{hyperref}
\usepackage[table,xcdraw]{xcolor}
\usepackage{subcaption}
\usepackage{graphicx}
\usepackage{colortbl}
\usepackage{doi}
\usepackage{caption}
\usepackage{hhline} % for horizontal lines in tables with background color in cells
\usepackage{authblk}
\usepackage{geometry}
\title{Cross-Modality Sub-Image Retrieval using Contrastive Multimodal Image Representations}

\author[1,+]{Eva Breznik}
\author[1,+]{Elisabeth Wetzer}
\author[1]{Joakim Lindblad}
\author[1]{Nata\v{s}a Sladoje}
\affil[1]{Uppsala University, Department of Information Technology, Uppsala, 751 05, Sweden}
\affil[+]{these authors contributed equally to this work}
\date{}
\begin{document}
\maketitle
\begin{abstract}
In tissue characterization and cancer diagnostics, multimodal imaging has emerged as a powerful technique. Thanks to computational advances, large datasets can be exploited to discover patterns in pathologies and improve diagnosis. However, this requires efficient and scalable image retrieval methods. Cross-modality image retrieval is particularly challenging, since images of similar (or even the same) content captured by different modalities might share few common structures. We propose a new application-independent content-based image retrieval (CBIR) system for reverse (sub-)image search across modalities, which combines deep learning to generate representations (embedding the different modalities in a common space) with classical feature extraction and bag-of-words models for efficient and reliable retrieval. 
We illustrate its advantages through a replacement study, exploring a number of feature extractors and learned representations, as well as through comparison to recent (cross-modality) CBIR methods. 
For the task of (sub-)image retrieval on a (publicly available) dataset of brightfield and second harmonic generation microscopy images,  the results show that our approach is superior to all tested alternatives.
We discuss the shortcomings of the compared methods and observe the importance of equivariance and invariance properties of the learned representations and feature extractors in the CBIR pipeline.
Code is available at: \url{https://github.com/MIDA-group/CrossModal_ImgRetrieval}.
\end{abstract}

%\flushbottom
%\maketitle
% * <john.hammersley@gmail.com> 2015-02-09T12:07:31.197Z:
%
%  Click the title above to edit the author information and abstract
%
%\thispagestyle{empty}

\section*{Introduction}
% CBIR GENERAL
Content-based image retrieval (CBIR) systems are designed to search images in large databases based on \textit{content}. Queries may be provided in various forms such as class labels, key words and images. The type of CBIR using images as queries is termed Reverse Image Search (RIS), %~\cite{TinEye},
also known as \textit{query-by-example}. CBIR systems traditionally consist of a feature extraction method followed by matching based on a suitable similarity measure~\cite{cbirs, sota}. Following the advent of deep learning, feature extraction is often performed by convolutional neural networks (CNNs), sometimes using pretrained networks~\cite{deepret, 9313211, 10.1007/978-3-030-89128-2_28, KongSRF_BMVC_2017}. 
When only a small patch of an image is provided as a query, the CBIR is termed sub-image retrieval (s-CBIR).

Often local feature descriptors are accumulated into a bag-of-words (BoW, also called bag-of-features)\cite{bag, bag2, 10.1007/978-3-642-02976-9_17}, where the most descriptive features (words) form a vocabulary and each image is assigned a histogram of words. The retrieval step is then based on histogram comparison, typically using cosine similarity.
In some approaches, global features are used in the first stage of image retrieval to find the most similar images within a dataset whereafter the top results are re-ranked using local features~\cite{10.1007/978-3-030-58565-5_43}. 

CBIR systems have gained popularity in digital pathology~\cite{smily, Komura345785, lora408237} due to the increased use of whole slide image (WSI) scanners which enable lowered storage costs of glass slides as well as simplified transportation of samples, training of new experts, spatial navigation of the sample~\cite{CHEN2020105630}, and, most importantly, powerful computer assisted sample analysis~\cite{nephro}. 
By supporting efficient searches through the huge datasets acquired, CBIR techniques facilitate diagnostic decision-making through easy access to similar previous cases and provide the potential to unravel patterns useful for early diagnosis of diseases such as cancer~\cite{nephro}, thereby enabling screening programs.

% MULTIMODAL
WSI scanners generally capture a single modality, usually fluorescent or brightfield (BF) microscopy. Acquiring additional images by different sensors may provide highly relevant complementary information. However, with the explicit aim to capture \textit{different} types of information, the acquired images may have very different appearances and share few structures. For many medical diagnoses, in particular cancer diagnosis and grading, manual examination of tissue samples with a hematoxylin and eosin (H\&E) stain using BF microscopy~\cite{Hristu:21} is the gold standard. 
In recent years, the label-free, non-linear imaging modality of second harmonic generation (SHG) proved to be useful for diagnostics for a variety of tissues, such as skin, ovaries and breast %, such as kidneys~\cite{Round1}, ovaries~\cite{doi:10.4137/PMC.S13214}, pancreas~\cite{Round2}, thyroid~\cite{Hristu:18}, breast~\cite{Eliceiri}%, Golaraei:16, doi:10.4137/PMC.S13214}
among others \cite{KEIKHOSRAVI2014531}. To facilitate content understanding, SHG images are often inspected side by side with corresponding BF images.
To fully exploit the advantages of such (large) multimodal datasets, the ability to query them across modalities can serve as a very important and useful tool to aid the diagnostic process. 
Furthermore, WSI scanners capture a large tissue area, often up to $100,000 \!\times\! 100,000$\,px, while SHG imaging at the same scale can typically only cover smaller areas. Hence SHG images can be taken at various locations within the tissue sample and provide local samples of additional information to a WSI BF image, emphasizing the importance of cross-modal methods for sub-image retrieval in particular.

%Furthermore, CBIR can also serve as a first step in image registration pipelines to find the images or regions which should be registered, since many registration methods are based on the assumption that %the identities of 
%the reference-target image pairs are known, which is not always the case in microscopy \cite{10.1007/978-3-031-11203-4_18}. This particularly holds for whole slide images, as WSI scanners capture a large tissue area, often up to 100,000 $\!\times\!$ 100,000\,px. In the same scale, SHG imaging can typically only cover smaller areas. Hence SHG images can be taken at various locations within the entire tissue sample and serve as local samples of additional information.

 \begin{figure*}
  \centering
  \includegraphics[width=\textwidth]{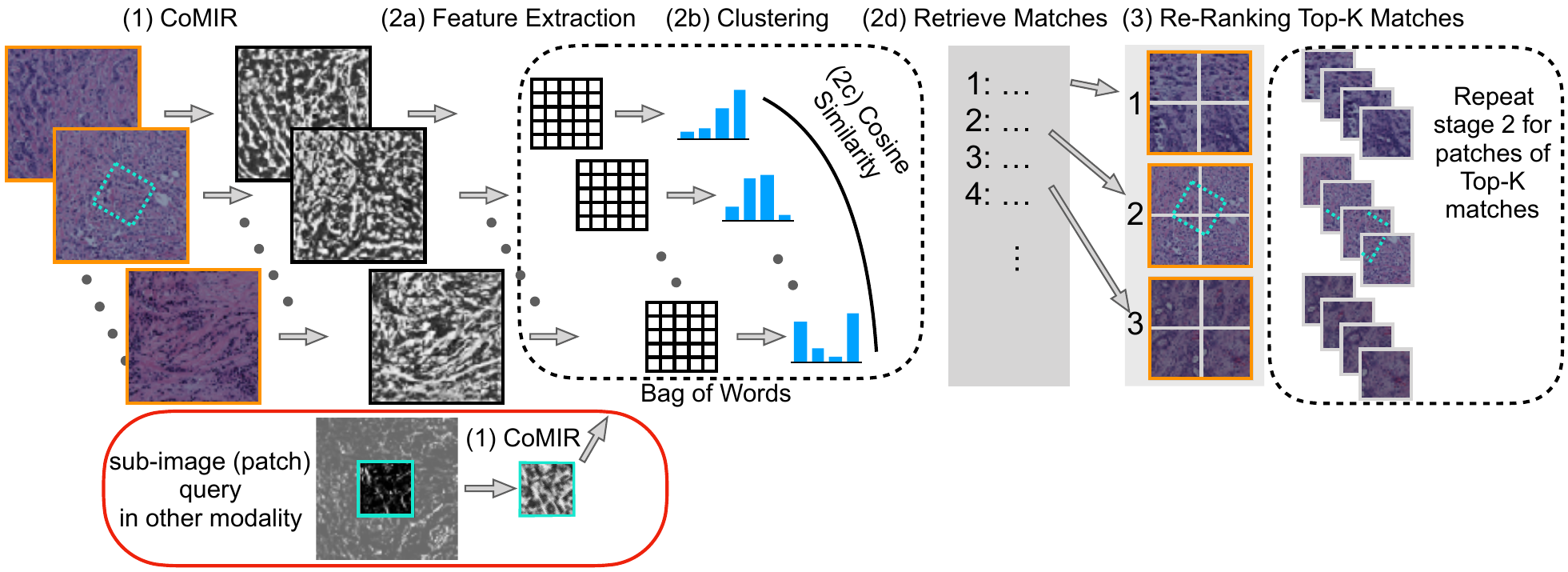}
  \caption[Structural overview of the proposed s-CBIR pipeline]{The proposed s-CBIR pipeline consists of three stages. Stage I includes learning CoMIRs for the images in the repository and the query (either as a fullsized image or in form of a patch), followed by sparse feature extraction in Stage II, which are binned into single descriptors for each image, building the vocabulary for a BoW. Matches are found using the cosine similarity. In Stage III, the Top-K matches are split into patches and a new BoW is computed for Re-Ranking.}
  \label{fig:1}
\end{figure*}

Cross-modality image retrieval (CMIR), i.e.  retrieval of images in one modality or visual domain provided a query in another, is also referred to as \textit{cross-domain image retrieval} (CDIR) or \textit{cross-source image retrieval}. A recent review on CDIR \cite{CrossDomainRev} addresses work done in the field of near infrared \& visible light retrieval used in person re-identification~\cite{9312626, 8237837}, synthetic aperture radar \& optical image as well as multispectral images \& panchromatic images retrieval in remote sensing~\cite{9044737, 8985543, 8385104}, and sketch \& natural image retrieval~\cite{tcnet,10.1007/978-3-030-01216-8_19, BAI2020102835, BUI201727, 10.1155/2021/5577735}. 
The methods used in these applications can be categorized into two types based on their approach to solving the domain gap problem~\cite{CrossDomainRev}: feature space migration~\cite{deepret, 9044737, 9312626, 9313211, 8385104, KongSRF_BMVC_2017} and image domain migration~\cite{BUI201727, 10.1007/978-3-030-01216-8_19, 8985543, BAI2020102835,10.1155/2021/5577735}. The first type extracts features of images in both modalities and attempts to find a mapping function (often through contrastive losses) to ensure feature similarity of corresponding image pairs or classes. The second relies on generative networks to translate one modality into the other ~\cite{10.1007/978-3-030-01216-8_19, 8985543, BAI2020102835, 10.1155/2021/5577735}.  
However, these methods often employ domain-specific steps (e.g. incorporation of edge information in sketch retrieval), or focus on category-level retrieval (assuming a query corresponds to more than one image in the repository, i.e. to all images of the same category) relying on class labels with multiple samples per class. Hence they are not applicable to very different domains such as microscopy images or instance-level retrieval (where each query has a single correct match in the repository) in general. Hashing based methods, which have been successfully used for image retrieval within medical imaging~\cite{9222290, FANG2021101981}, have also been applied to cross-modality retrieval\cite{9222290,9313211}, but they generally use class labels for training, assuming multiple instances per class. 
 
%Hashing based methods have been successfully used for image retrieval within various medical image modalities %for creating lower dimensional representation useful in image retrieval 
%\cite{9222290, FANG2021101981}. While they can be applicable also in cross-modality retrieval\cite{9222290,9313211}, they generally use class labels for training, assuming multiple instances per class. 
%However, these methods often employ domain-specific steps (e.g. incorporation of edge information in sketch retrieval), large training datasets, or are specific to category-level retrieval (assuming a query corresponds to multiple images in the repository, i.e. to all images of the same category/class) and are hence not applicable to our task in which we perform instance-level retrieval (i.e. each query image corresponds to a single correct match in the repository) on visually very different modalities for which training set labels (in form of image pair correspondences) are expensive.

%for mono- and cross-modal retrieval, and their combination with classical ones\cite{10.1007/978-3-030-89128-2_28} for magnetic resonance imaging (MRI) and computed tomography (CT) image retrieval. Hashing based methods have been successfully used for image retrieval within various medical image modalities %for creating lower dimensional representation useful in image retrieval 
%\cite{9222290, FANG2021101981}. While they can be applicable also in cross-modality retrieval\cite{9222290,9313211}, they generally use class labels for training, assuming multiple instances per class.  

In this paper, we study cross-modality image retrieval formulated as %reverse image search 
RIS on an instance level, i.e. the images are unlabelled and the aim is to retrieve the image of modality \textit{A} corresponding to a given query image of modality \textit{B}. 
We propose a data-independent three-stage s-CBIR system that  
uses representation learning to transform images of both modalities into a common space via contrastive multimodal image representations for registration (CoMIRs)~\cite{pielawski2020comir}, followed by feature extraction and a BoW model to perform retrieval in this abstract representation space. Finally, we suggest re-ranking the top results to further improve the performance.

% WHAT WE DO + CONTRIBUTIONS
The proposed approach is evaluated on the very challenging task of (sub-)image retrieval across the BF and SHG modalities, since these two modalities are too different in their appearances to enable successful retrieval using existing monomodal RIS approaches. The evaluation is performed through a replacement study, using a number of viable alternatives for each step of the proposed pipeline. In addition, we compare our method to recent state-of-the-art methods in cross-modality and biomedical image retrieval.

\textbf{Contributions:} We propose a state-of-the-art cross-modality sub-image retrieval system for reverse image search which combines CoMIR representation learning and SURF feature extraction. We carry out a replacement study to demonstrate its efficacy. Our proposed approach outperforms state-of-the-art methods on the challenging task of image retrieval across BF an SHG modalities.
Furthermore, we: (i) discuss the shortcomings of the I2I based approaches and highlight the necessity of rotationally equivariant representations for translating the multimodal task into a monomodal one; (ii) demonstrate the importance of rotational invariance of the feature extractor; and (iii) show that re-ranking can boost the retrieval performance significantly. We share the code as open source at \url{https://github.com/MIDA-group/CrossModal_ImgRetrieval}.

\section*{Method}
\label{sec:met}

Our aim is to match corresponding areas of two different modalities which may be hard to align even by human inspection~\cite{pielawski2020comir}. %, as for example in the case of BF and SHG~\cite{pielawski2020comir}. 
The proposed pipeline is modular, with three main stages. The first stage uses representation learning to bridge between the modalities, the second stage consists of feature extraction, bag of words computation and match retrieval, followed by re-ranking in the third stage, in which a new BoW of the top retrieval results is computed, as depicted in Figure \ref{fig:1}.

\subsection*{Stage I: Representation Learning}
We use representation learning to bridge the gap between the input modalities inspired by the success of other recent multimodal image retrieval approaches which rely on domain migration~\cite{BUI201727, 10.1007/978-3-030-01216-8_19, 8985543, BAI2020102835,10.1155/2021/5577735}. Our proposed pipeline does so by using contrastive learning (step (1) in Fig. \ref{fig:1}). 
Contrastive losses are used in a number of multimodal image retrieval tasks to learn feature embeddings which are similar for corresponding samples~\cite{9044737,BUI201727}. In the proposed method we use \textit{contrastive multimodal image representations (called CoMIRs)}, which are image representations learned by training two CNNs in parallel with aligned image pairs of different modalities.  
Using a contrastive loss, the two networks produce representations of the input images, such that two CoMIRs resulting from corresponding areas in the two input modalities have maximum similarity w.r.t.\ a selected similarity measure. The networks are provided with randomly chosen $\{0^{\circ},90^{\circ},180^{\circ}, 270^{\circ}\}$-rotated versions of the input images, which are aligned with the corresponding input of the other modality in the second network before the contrastive loss is computed, thereby enforcing rotationally equivariant properties of the representations. The representations preserve common structures, which makes them useful for multimodal image registration and hence suitable candidates for image retrieval. For more details on the method and implementation see Pielawski \& Wetzer et al.~\cite{pielawski2020comir}.

\subsection*{Stage II: Feature Extraction and Creation of BoW}
\label{ssec:scbir}
This stage consists of extracting features from the CoMIR images and building a BoW model based on them. 
We employ Speeded Up Robust Features (SURF)~\cite{SURF} (step (2a) in Fig.~\ref{fig:1}) which are sparse, scale- and rotation invariant, hence they are expected to perform well even with rigidly transformed or cropped queries.
The BoW is defined on the features extracted from the CoMIRs of all the images in the searchable repository, by K-means clustering using a suitable vocabulary (step (2b) in Fig.~\ref{fig:1}). The features extracted from the CoMIR of the (rigidly transformed) query image are encoded using the created vocabulary, resulting in a histogram of features associated to the BoW. This histogram is then matched against the database using cosine similarity (steps (2c\&2d) in Fig. \ref{fig:1}) to retrieve the best matches.

\subsection*{Stage III: Re-ranking}
To further improve the retrieval of our (s\nobreakdash-)CBIR system, the best ranked matches can be re-ranked. To do so, we take a number of top retrieval matches and cut them into patches of the same size as the query, as shown in step (3) in Fig.~\ref{fig:1}. The resulting patches of the top matches form a database for which a new BoW model and (s\nobreakdash-)CBIR ranking is computed, using the same configuration as the initial one (step (3) in Fig.~\ref{fig:1}). In case of full-image search no cutting is performed. Instead the new ranking is computed directly on the set of the top ranked matches.

\section*{Evaluation}
\label{sec:exp}
The aim is to retrieve a (transformed) query (sub-)image from a repository storing the other modality. A successful match is defined as the retrieval of the image corresponding to the query of the other modality (instance-level retrieval). The evaluation is done exhaustively, using all images in the (one-modality) dataset as queries. A top-K retrieval success indicates for what fraction of queries a correct image was found in the first K matches (often denoted by Acc@k). 
To thoroughly evaluate the proposed method we run a replacement study on its individual modules, as well as compare it against current state-of-the-art in CMIR. The evaluation is performed for both full-sized and small patch queries.

\subsection*{Dataset}
\label{ssec:data}
\begin{figure}
  \centering
  \includegraphics[width=0.6\textwidth]{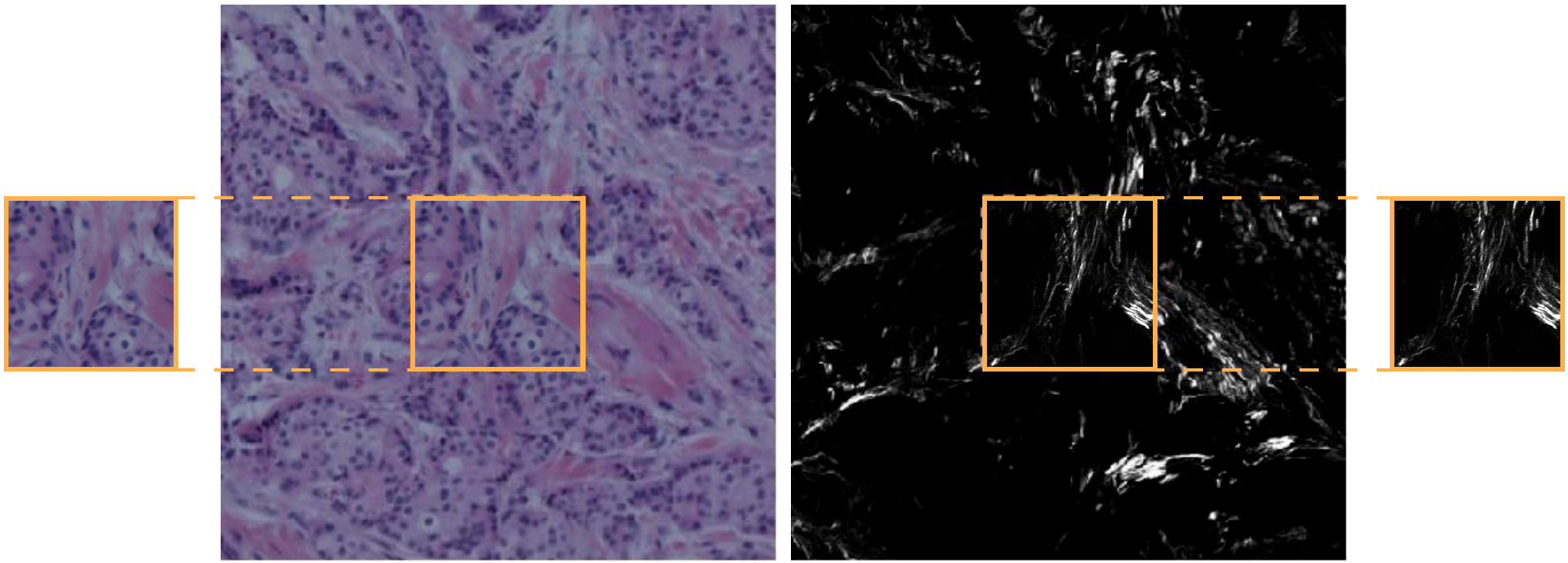}
  \caption[Example images of the data set used in the experiments]{Example of a BF (left) and SHG (right) image pair used in the CBIR experiments. The orange squares indicate the patches cropped from the images used in the s-CBIR experiments. This shown image pair is aligned, without the rotations and translations used in the test set.}
  \label{fig:2}
\end{figure}

Our evaluation dataset consists of $206$ BF and SHG image pairs of size $834 \times 834$\,px, and is an openly available registration benchmark \cite{kevin_eliceiri_2020_3874362}. %, originating from Eliceiri et al.~\cite{fullcores}.
For each image pair, also its rigidly transformed version is provided. The transformations consist of a random rotation up to $\pm30^{\circ}$, and random translations up to $\pm100$\,px in $x$ and $y$. 
Following Pielawski \& Wetzer et al.~\cite{pielawski2020comir}, $40$ untransformed pairs are used for training, $134$ for testing, and the remaining $32$ for validation. SHG images were preprocessed by a log-transform for the I2I and CoMIR generation. For evaluating the s-CBIR,  patches of size $256\times 256$\,px are cropped from the centres of the $834\times834$\,px images. 
An example pair is shown in Fig. \ref{fig:2}. 

\subsection*{Experimental settings for the proposed method}
%\textbf{CoMIRs}: 
To learn 1-channel CoMIRs, two U-Nets~\cite{jegou2017one} sharing no weights are used under the same settings as in Pielawski \& Wetzer et al.~\cite{pielawski2020comir} for registration, with mean squared error as a critic, manually tuned temperature $\tau=0.5$ and 46 negative pairs in each iteration. The networks are trained on patches of size $128 \times 128$\,px randomly extracted from 40 aligned training pairs. 
%\textbf{SURF}:
The SURF features are extracted for patch sizes $32, 64, 96,$ and $128$ on a grid with spacing $(8,8)$, using a descriptor size of 64.
The BoW is then defined on the features of the entire set of untransformed CoMIRs, using a large vocabulary of 20,000 words (based on empirical testing) on the $80\%$ of the strongest features and the matching is done with cosine similarity.
%\textbf{Re-Ranking}: 
We do the final re-ranking among top-15 and top-30 matches. For s-CBIR, the image is cut into the minimal number of equidistantly placed query-sized patches s.t. the image is fully covered.

\subsection*{Replacement study}
\label{ssec:exp}
A replacement study is performed on stages I and II of the pipeline. 
Re-ranking is performed at the end, only where base results suggest its applicability (i.e. only within the proposed pipeline).  

\subsubsection*{Stage I alternatives} For bridging the gap between modalities, CoMIR representations are compared against two generative adversarial network (GAN)-based image-to-image translation (I2I)~\cite{isola2017image, CycleGAN2017} methods, used to translate BF images into SHG images, and vice versa.\\ 
GANs consist of a generator and a discriminator, competing in a zero-sum game. The generator learns to generate a representation in one modality given the input in the other modality; the discriminator classifies the representation as generated or real, thereby training the generator to produce representations indistinguishable from real images. The resulting so-called "fake" BF and SHG images can be used as queries, enabling search in near-monomodality. %within their respective target modality.  
\\
\textbf{Pix2pix}~\cite{isola2017image} uses a conditional GAN to learn a mapping from an input image to a representation using corresponding images of both modalities, i.e. requires supervision in form of aligned multimodal image pairs. \textbf{CycleGAN}~\cite{CycleGAN2017} differs from pix2pix by achieving this goal even in an unsupervised manner, through a cycle consistency enforcing that the input image can be reconstructed  
from the representation that was produced based on it. CycleGAN-based domain adaptation has been used successfully in previous multimodal image retrieval tasks~\cite{8985543, 10.1155/2021/5577735}.\\
For both approaches, training parameters are used as in Pielawski \& Wetzer et al.~\cite{pielawski2020comir}. The code is available at \url{https://github.com/junyanz/pytorch-CycleGAN-and-pix2pix}.

\subsubsection*{Stage II alternatives}
For the second stage we compare the choice of SURF against two other commonly used feature extractors,  SIFT~\cite{790410} and ResNet~\cite{7780459}. Regardless of the feature extractor choice, the BoW is defined on the features of the entire set of untransformed input images of one modality (or their learned representations such as CoMIRs), using a vocabulary of 20,000 words on the $80\%$ of the strongest features and the matching is done with cosine similarity.
In addition, we test replacing the entire stage II (feature extraction and BoW) with a recent RIS s-CBIR toolkit 2KDK~\cite{2KDK}, based on 2D Krawtchouk descriptors.\\ 
\textbf{SIFT} (Scale Invariant Feature Transform)~\cite{790410} is a well known feature extractor with similar properties as SURF, being sparse, scale and rotation invariant. In our evaluation the feature descriptor size is 4 samples per row and column, 8 bins per local histogram. The range of the scale octaves is $[32,512]$\,px, with 4 steps per scale octave and an initial $\sigma$ of each scale octave equal to $1.6$. The descriptor size is 128.
We explore \textbf{ResNet}~\cite{7780459} as a dense feature extractor based on previous successes of using features extracted by neural networks for image retrieval~\cite{babenko2015, satelit2019, DBLP:journals/corr/abs-1903-10663, lora408237}. As data used in this study is not required to have labels, we extract features by ResNet152, pretrained on ImageNet\cite{DBLP:journals/corr/abs-1903-10663} (see  \url{https://pytorch.org/vision/stable/models.html}). 
The features are extracted by removing the last fully connected layer. To enable patch queries, an adaptive average pooling is added to produce features of size $8\times8$ (64 when flattened), independent of the input size. The number of extracted features is 2048 regardless of the input image size. For SHG, the image is copied into three channels. 
As opposed to using SIFT/SURF which extract an arbitrary number of features from every image, the amount of ResNet extracted features is the same for every image (or patch).
\textbf{2DKD Toolkit}~\cite{2KDK} is a recent RIS system which differs from the BoW approach in that it is performing a local sub-image search using a number of translation, rotation, and scaling invariant descriptors per image. It relies on moment invariants based on Krawtchouk polynomials~\cite{6783793}, namely Two Dimensional Krawtchouk Descriptors, which outperform Hu invariants in retrieving subimages in cryo-electron microscopy images in a monomodal setting~\cite{2KDK}. The authors point out the importance of using moment invariants that do not change by translation, rotation and scaling in digital pathology.
In our evaluation, the number of pixels between two consecutive points of interest is set to 5 and local pixel intensity variance is used as a criterion to compare against global pixel density variance. The general experimental setup follows the one of DeVille et al.~\cite{2KDK}.

\subsection*{Competing Methods}
Since methods for instance-level cross-modality retrieval (specifically applicable in the biomedical field) are scarce, we compare our method to (1) a recent CMIR method, Triplet Classification Network\cite{tcnet} (TC-Net) which was developed primarily for retrieval across sketches and natural images but is not domain-specific in design and can be used for instance-level retrieval, and (2) a medical image retrieval system based on invariant moments, textural and deep features~\cite{10.1007/978-3-030-89128-2_28} (IMTDF), which has been developed for, and evaluated on, within-modality retrieval, but relies in part on a study suggesting cross-modal applicability~\cite{9313211}. While more recent CMIR methods exist, they are not applicable for our particular evaluation task. %specific evaluation problem case. 
\\
\textbf{TC-Net} improves on the previous works\cite{dssa} in cross-modality sketch retrieval by circumventing the generation of edge maps, as their quality has a large effect on CBIR system performance. It uses a triplet Siamese network, and auxiliary classification losses. For the evaluation, we use the settings of the original paper\cite{tcnet}, however for a fairer comparison we train the network with BF anchors for retrieval within SHG database, and with SHG anchors for retrieval within BF database.
\\
\textbf{IMTDF} relies on a combination of various invariant moments, classical texture features and CNN based features. It thereby follows other RIS methods in biomedical applications which rely on CNN features~\cite{deepret, 9313211} for both mono- and cross-modal retrieval, and their combination with classical features, as successfully used for cross-modal retrieval of magnetic resonance imaging (MRI) and computed tomography (CT) images~\cite{10.1007/978-3-030-89128-2_28}.
Based on the best performing features reported for IMTDF~\cite{10.1007/978-3-030-89128-2_28}, we use a combination of Chebyshev moments, Haralick texture features (converted to rotationally invariant), and ResNet50 features, selecting only the  strongest $20\%$ of the latter two by means of ReliefF. For more details and parameter settings see the original paper\cite{10.1007/978-3-030-89128-2_28}.

\begin{table}[t]
%\resizebox{0.49\textwidth}{!}{%
\begin{subtable}[c]{0.49\textwidth}
\centering
\resizebox{\textwidth}{!}{
\begin{tabular}{clllrrrr}
 & & & &
  \multicolumn{4}{c}{\textbf{Query}} \\ \cline{2-8} 
\multicolumn{1}{c|}{} &
  \multicolumn{1}{l|}{} &
  \multicolumn{1}{l|}{} &
  \multicolumn{1}{l|}{} &
  \multicolumn{1}{l|}{BF} &
  \multicolumn{1}{l|}{BF(T)} &
  \multicolumn{1}{l|}{SHG} &
  \multicolumn{1}{l|}{SHG(T)} \\ \cline{4-8} 
\multicolumn{1}{c|}{} &
  \multicolumn{1}{l|}{} &
  \multicolumn{1}{l|}{} &
  \multicolumn{1}{l|}{BF} &
  \cellcolor[HTML]{A1CEF0}100.0 &
  \cellcolor[HTML]{FFD570}100.0 &
  9.7 &
  \multicolumn{1}{r|}{\cellcolor[HTML]{FF9CA2}11.2} \\ \cline{4-4}
\multicolumn{1}{c|}{} &
  \multicolumn{1}{l|}{} &
  \multicolumn{1}{l|}{\multirow{-3}{*}{Originals}} &
  \multicolumn{1}{l|}{SHG} &
  9.2 &
  \cellcolor[HTML]{FF9CA2}10 &
  \cellcolor[HTML]{A1CEF0}100.0 &
  \multicolumn{1}{r|}{\cellcolor[HTML]{FFD570}100.0} \\ \cline{3-8} 
\multicolumn{1}{c|}{} &
  \multicolumn{1}{l|}{} &
  \multicolumn{1}{l|}{} &
  \multicolumn{1}{l|}{} &
  \multicolumn{1}{l|}{CoMIR(BF)} &
  \multicolumn{1}{l|}{CoMIR(BF(T))} &
  \multicolumn{1}{l|}{CoMIR(SHG)} &
  \multicolumn{1}{l|}{CoMIR(SHG(T))} \\ \cline{4-8} 
\multicolumn{1}{c|}{} &
  \multicolumn{1}{l|}{} &
  \multicolumn{1}{l|}{} &
  \multicolumn{1}{l|}{CoMIR(BF)} &
  \cellcolor[HTML]{A1CEF0}100.0 &
  \cellcolor[HTML]{FFD570}97 &
  37.1 &
  \multicolumn{1}{r|}{{\cellcolor[HTML]{FF9CA2} 32.8}} \\ \cline{4-4}
\multicolumn{1}{c|}{} &
  \multicolumn{1}{l|}{} &
  \multicolumn{1}{l|}{\multirow{-3}{*}{CoMIR}} &
  \multicolumn{1}{l|}{CoMIR(SHG)} &
  31.1 &
  \cellcolor[HTML]{FF9CA2}30.3 &
  \cellcolor[HTML]{A1CEF0}100.0 &
  \multicolumn{1}{r|}{\cellcolor[HTML]{FFD570}100.0} \\ \cline{3-8} 
\multicolumn{1}{c|}{} &
  \multicolumn{1}{l|}{} &
  \multicolumn{1}{l|}{} &
  \multicolumn{1}{l|}{} &
  \multicolumn{1}{l|}{BF} &
  \multicolumn{1}{l|}{BF(T)} &
  \multicolumn{1}{l|}{Fake BF} &
  \multicolumn{1}{l|}{Fake BF(T)} \\ \cline{4-8} 
\multicolumn{1}{c|}{} &
  \multicolumn{1}{l|}{} &
  \multicolumn{1}{l|}{} &
  \multicolumn{1}{l|}{BF} &
 \cellcolor[HTML]{A1CEF0} 100.0 &
  {\cellcolor[HTML]{FFD570}100.0} &
  15.7 &
  \multicolumn{1}{r|}{\cellcolor[HTML]{FF9CA2}16.4} \\ \cline{4-4}
\multicolumn{1}{c|}{} &
  \multicolumn{1}{l|}{} &
  \multicolumn{1}{l|}{} &
  \multicolumn{1}{l|}{Fake BF} &
  26.1 &
  \cellcolor[HTML]{FF9CA2}24.6 &
  \cellcolor[HTML]{A1CEF0}100.0 &
  \multicolumn{1}{r|}{\cellcolor[HTML]{FFD570}70.9} \\ \cline{4-8} 
\multicolumn{1}{c|}{} &
  \multicolumn{1}{l|}{} &
  \multicolumn{1}{l|}{} &
  \multicolumn{1}{l|}{} &
  \multicolumn{1}{l|}{SHG} &
  \multicolumn{1}{l|}{SHG(T)} &
  \multicolumn{1}{l|}{Fake SHG} &
  \multicolumn{1}{l|}{Fake SHG(T)} \\ \cline{4-8} 
\multicolumn{1}{c|}{} &
  \multicolumn{1}{l|}{} &
  \multicolumn{1}{l|}{} &
  \multicolumn{1}{l|}{SHG} &
  \cellcolor[HTML]{A1CEF0}100.0 &
  {\cellcolor[HTML]{FFD570} 100.0 } &
  14.2 &
  \multicolumn{1}{r|}{\cellcolor[HTML]{FF9CA2}13.4} \\ \cline{4-4}
\multicolumn{1}{c|}{} &
  \multicolumn{1}{l|}{} &
  \multicolumn{1}{l|}{\multirow{-6}{*}{CycleGAN}} &
  \multicolumn{1}{l|}{Fake SHG} &
  15.7 &
  \cellcolor[HTML]{FF9CA2}18.7 &
  \cellcolor[HTML]{A1CEF0}100.0 &
  \multicolumn{1}{r|}{\cellcolor[HTML]{FFD570}48.5} \\ \cline{3-8} 
\multicolumn{1}{c|}{} &
  \multicolumn{1}{l|}{} &
  \multicolumn{1}{l|}{} &
  \multicolumn{1}{l|}{} &
  \multicolumn{1}{l|}{BF} &
  \multicolumn{1}{l|}{BF(T)} &
  \multicolumn{1}{l|}{Fake BF} &
  \multicolumn{1}{l|}{Fake BF(T)} \\ \cline{4-8} 
\multicolumn{1}{c|}{} &
  \multicolumn{1}{l|}{} &
  \multicolumn{1}{l|}{} &
  \multicolumn{1}{l|}{BF} &
  \cellcolor[HTML]{A1CEF0}100.0 &
  {\cellcolor[HTML]{FFD570}100.0} &
  14.2 &
  \multicolumn{1}{r|}{\cellcolor[HTML]{FF9CA2}21.6} \\ \cline{4-4}
\multicolumn{1}{c|}{} &
  \multicolumn{1}{l|}{} &
  \multicolumn{1}{l|}{} &
  \multicolumn{1}{l|}{Fake BF} &
  20.9 &
  \cellcolor[HTML]{FF9CA2}22.4 &
  \cellcolor[HTML]{A1CEF0}100.0 &
  \multicolumn{1}{r|}{\cellcolor[HTML]{FFD570}20.2} \\ \cline{4-8} 
\multicolumn{1}{c|}{} &
  \multicolumn{1}{l|}{} &
  \multicolumn{1}{l|}{} &
  \multicolumn{1}{l|}{} &
  \multicolumn{1}{l|}{SHG} &
  \multicolumn{1}{l|}{SHG(T)} &
  \multicolumn{1}{l|}{Fake SHG} &
  \multicolumn{1}{l|}{Fake SHG(T)} \\ \cline{4-8} 
\multicolumn{1}{c|}{} &
  \multicolumn{1}{l|}{} &
  \multicolumn{1}{l|}{} &
  \multicolumn{1}{l|}{SHG} &
 \cellcolor[HTML]{A1CEF0} 100.0 &
  {\cellcolor[HTML]{FFD570}100.0 }&
  14.2 &
  \multicolumn{1}{r|}{\cellcolor[HTML]{FF9CA2}14.2} \\ \cline{4-4}
\multicolumn{1}{c|}{} &
  \multicolumn{1}{l|}{\multirow{-18}{*}{\rotatebox[origin=c]{90}{SIFT}}} &
  \multicolumn{1}{l|}{\multirow{-6}{*}{Pix2Pix}} &
  \multicolumn{1}{l|}{Fake SHG} &
  0.0 &
  \cellcolor[HTML]{FF9CA2}18.7 &
  \cellcolor[HTML]{A1CEF0}100.0 &
  \multicolumn{1}{r|}{\cellcolor[HTML]{FFD570}26.1} \\ \cline{2-8} \noalign{\vskip\doublerulesep\vskip-\arrayrulewidth} \cline{2-8}
\multicolumn{1}{c|}{} &
  \multicolumn{1}{l|}{} &
  \multicolumn{1}{l|}{} &
  \multicolumn{1}{l|}{} &
  \multicolumn{1}{l|}{BF} &
  \multicolumn{1}{l|}{BF(T)} &
  \multicolumn{1}{l|}{SHG} &
  \multicolumn{1}{l|}{SHG(T)} \\ \cline{4-8} 
\multicolumn{1}{c|}{} &
  \multicolumn{1}{l|}{} &
  \multicolumn{1}{l|}{} &
  \multicolumn{1}{l|}{BF} &
  \cellcolor[HTML]{A1CEF0}100.0 &
  \cellcolor[HTML]{FFD570} 97.3 &
  15.2 &
  \multicolumn{1}{r|}{\cellcolor[HTML]{FF9CA2}14.2} \\ \cline{4-4}
\multicolumn{1}{c|}{} &
  \multicolumn{1}{l|}{} &
  \multicolumn{1}{l|}{\multirow{-3}{*}{Originals}} &
  \multicolumn{1}{l|}{SHG} &
  10.9 &
  \cellcolor[HTML]{FF9CA2}10.7 &
  \cellcolor[HTML]{A1CEF0}100.0 &
  \multicolumn{1}{r|}{\cellcolor[HTML]{FFD570}97.5} \\ \cline{3-8} 
\multicolumn{1}{c|}{} &
  \multicolumn{1}{l|}{} &
  \multicolumn{1}{l|}{} &
  \multicolumn{1}{l|}{} &
  \multicolumn{1}{l|}{CoMIR(BF)} &
  \multicolumn{1}{l|}{CoMIR(BF(T))} &
  \multicolumn{1}{l|}{CoMIR(SHG)} &
  \multicolumn{1}{l|}{CoMIR(SHG(T))} \\ \cline{4-8} 
\multicolumn{1}{c|}{} &
  \multicolumn{1}{l|}{} &
  \multicolumn{1}{l|}{} &
  \multicolumn{1}{l|}{CoMIR(BF)} &
  \cellcolor[HTML]{A1CEF0}100.0 &
  \cellcolor[HTML]{FFD570} 93.8 &
  71.4 &
  \multicolumn{1}{r|}{\cellcolor[HTML]{FF9CA2}\textbf{61.2}} \\ \cline{4-4}
\multicolumn{1}{c|}{} &
  \multicolumn{1}{l|}{} &
  \multicolumn{1}{l|}{\multirow{-3}{*}{CoMIR}} &
  \multicolumn{1}{l|}{CoMIR(SHG)} &
  76.1 &
  \cellcolor[HTML]{FF9CA2}\textbf{59.0} &
  \cellcolor[HTML]{A1CEF0}100.0 &
  \multicolumn{1}{r|}{\cellcolor[HTML]{FFD570}92.1} \\ \cline{3-8} 
\multicolumn{1}{c|}{} &
  \multicolumn{1}{l|}{} &
  \multicolumn{1}{l|}{} &
  \multicolumn{1}{l|}{} &
  \multicolumn{1}{l|}{BF} &
  \multicolumn{1}{l|}{BF(T)} &
  \multicolumn{1}{l|}{Fake BF} &
  \multicolumn{1}{l|}{Fake BF(T)} \\ \cline{4-8} 
\multicolumn{1}{c|}{} &
  \multicolumn{1}{l|}{} &
  \multicolumn{1}{l|}{} &
  \multicolumn{1}{l|}{BF} &
 \cellcolor[HTML]{A1CEF0} 100.0 &
  {\cellcolor[HTML]{FFD570}98.9 }&
  21.6 &
  \multicolumn{1}{r|}{\cellcolor[HTML]{FF9CA2}17.2} \\ \cline{4-4}
\multicolumn{1}{c|}{} &
  \multicolumn{1}{l|}{} &
  \multicolumn{1}{l|}{} &
  \multicolumn{1}{l|}{Fake BF} &
  35.8 &
  \cellcolor[HTML]{FF9CA2}23.9 &
  \cellcolor[HTML]{A1CEF0}100.0 &
  \multicolumn{1}{r|}{\cellcolor[HTML]{FFD570}78.4} \\ \cline{4-8} 
\multicolumn{1}{c|}{} &
  \multicolumn{1}{l|}{} &
  \multicolumn{1}{l|}{} &
  \multicolumn{1}{l|}{} &
  \multicolumn{1}{l|}{SHG} &
  \multicolumn{1}{l|}{SHG(T)} &
  \multicolumn{1}{l|}{Fake SHG} &
  \multicolumn{1}{l|}{Fake SHG(T)} \\ \cline{4-8} 
\multicolumn{1}{c|}{} &
  \multicolumn{1}{l|}{} &
  \multicolumn{1}{l|}{} &
  \multicolumn{1}{l|}{SHG} &
 \cellcolor[HTML]{A1CEF0} 100.0 &
  {\cellcolor[HTML]{FFD570}98.1}&
  12.7 &
  \multicolumn{1}{r|}{\cellcolor[HTML]{FF9CA2}12.7} \\ \cline{4-4}
\multicolumn{1}{c|}{} &
  \multicolumn{1}{l|}{} &
  \multicolumn{1}{l|}{\multirow{-6}{*}{CycleGAN}} &
  \multicolumn{1}{l|}{Fake SHG} &
  14.2 &
  \cellcolor[HTML]{FF9CA2}12.7 &
  \cellcolor[HTML]{A1CEF0}100.0 &
  \multicolumn{1}{r|}{\cellcolor[HTML]{FFD570}88.8} \\ \cline{3-8} 
\multicolumn{1}{c|}{} &
  \multicolumn{1}{l|}{} &
  \multicolumn{1}{l|}{} &
  \multicolumn{1}{l|}{} &
  \multicolumn{1}{l|}{BF} &
  \multicolumn{1}{l|}{BF(T)} &
  \multicolumn{1}{l|}{Fake BF} &
  \multicolumn{1}{l|}{Fake BF(T)} \\ \cline{4-8} 
\multicolumn{1}{c|}{} &
  \multicolumn{1}{l|}{} &
  \multicolumn{1}{l|}{} &
  \multicolumn{1}{l|}{BF} &
 \cellcolor[HTML]{A1CEF0} 100.0 &
  {\cellcolor[HTML]{FFD570}98.9} &
  20.9 &
  \multicolumn{1}{r|}{\cellcolor[HTML]{FF9CA2}12.7} \\ \cline{4-4}
\multicolumn{1}{c|}{} &
  \multicolumn{1}{l|}{} &
  \multicolumn{1}{l|}{} &
  \multicolumn{1}{l|}{Fake BF} &
  35.8 &
  \cellcolor[HTML]{FF9CA2}31.3 &
  \cellcolor[HTML]{A1CEF0}100.0 &
  \multicolumn{1}{r|}{\cellcolor[HTML]{FFDD8B}67.9} \\ \cline{4-8} 
\multicolumn{1}{c|}{} &
  \multicolumn{1}{l|}{} &
  \multicolumn{1}{l|}{} &
  \multicolumn{1}{l|}{} &
  \multicolumn{1}{l|}{SHG} &
  \multicolumn{1}{l|}{SHG (T)} &
  \multicolumn{1}{l|}{Fake SHG} &
  \multicolumn{1}{l|}{Fake SHG(T)} \\ \cline{4-8} 
\multicolumn{1}{c|}{} &
  \multicolumn{1}{l|}{} &
  \multicolumn{1}{l|}{} &
  \multicolumn{1}{l|}{SHG} &
  \cellcolor[HTML]{A1CEF0} 100.0 &
 {\cellcolor[HTML]{FFD570} 98.1 }&
  26.1 &
  \multicolumn{1}{r|}{\cellcolor[HTML]{FF9CA2}19.4} \\ \cline{4-4}
\multicolumn{1}{c|}{} &
  \multicolumn{1}{l|}{\multirow{-18}{*}{\rotatebox[origin=c]{90}{SURF}}} &
  \multicolumn{1}{l|}{\multirow{-6}{*}{Pix2Pix}} &
  \multicolumn{1}{l|}{Fake SHG} &
  35.1 &
  \cellcolor[HTML]{FF9CA2}31.3 &
  \cellcolor[HTML]{A1CEF0}100.0 &
  \multicolumn{1}{r|}{\cellcolor[HTML]{FFD570}85.1} \\ \cline{2-8} \noalign{\vskip\doublerulesep\vskip-\arrayrulewidth} \cline{2-8}
\multicolumn{1}{c|}{} &
  \multicolumn{1}{l|}{} &
  \multicolumn{1}{l|}{} &
  \multicolumn{1}{l|}{} &
  \multicolumn{1}{l|}{BF} &
  \multicolumn{1}{l|}{BF(T)} &
  \multicolumn{1}{l|}{SHG} &
  \multicolumn{1}{l|}{SHG(T)} \\ \cline{4-8} 
\multicolumn{1}{c|}{} &
  \multicolumn{1}{l|}{} &
  \multicolumn{1}{l|}{} &
  \multicolumn{1}{l|}{BF} &
  \cellcolor[HTML]{A1CEF0}100.0 &
  \cellcolor[HTML]{FFD570} 73.1 &
  48.5 &
  \multicolumn{1}{r|}{\cellcolor[HTML]{FF9CA2}29.1} \\ \cline{4-4}
\multicolumn{1}{c|}{} &
  \multicolumn{1}{l|}{} &
  \multicolumn{1}{l|}{\multirow{-3}{*}{Originals}} &
  \multicolumn{1}{l|}{SHG} &
  46.3 &
  \cellcolor[HTML]{FF9CA2}23.1 &
  \cellcolor[HTML]{A1CEF0}100.0 &
  \multicolumn{1}{r|}{\cellcolor[HTML]{FFD570}70.1} \\ \cline{3-8} 
\multicolumn{1}{c|}{} &
  \multicolumn{1}{l|}{} &
  \multicolumn{1}{l|}{} &
  \multicolumn{1}{l|}{} &
  \multicolumn{1}{l|}{CoMIR(BF)} &
  \multicolumn{1}{l|}{CoMIR(BF(T))} &
  \multicolumn{1}{l|}{CoMIR(SHG)} &
  \multicolumn{1}{l|}{CoMIR(SHG(T))} \\ \cline{4-8} 
\multicolumn{1}{c|}{} &
  \multicolumn{1}{l|}{} &
  \multicolumn{1}{l|}{} &
  \multicolumn{1}{l|}{CoMIR(BF)} &
  \cellcolor[HTML]{A1CEF0} 100.0 &
   \cellcolor[HTML]{FFD570} 61.6 & 50.8 &
  \multicolumn{1}{r|}{\cellcolor[HTML]{FF9CA2} 20.9} \\ \cline{4-4}
\multicolumn{1}{c|}{\multirow{-42}{*}{\rotatebox[origin=c]{90}{\textbf{Searchable repository}}}} &
  \multicolumn{1}{l|}{\multirow{-6}{*}{\rotatebox[origin=c]{90}{ResNet}}} &
  \multicolumn{1}{l|}{\multirow{-3}{*}{CoMIR}} &
  \multicolumn{1}{l|}{CoMIR(SHG)} &
   50.4 &
  \cellcolor[HTML]{FF9CA2} 22.7 &
  \cellcolor[HTML]{A1CEF0} 100.0 &
  \multicolumn{1}{r|}{\cellcolor[HTML]{FFD570} 53.7 } \\ \cline{2-8} \noalign{\vskip\doublerulesep\vskip-\arrayrulewidth} \cline{2-8} % BELOW HERE IS THE 2DKD SECTION
\multicolumn{1}{c|}{} &
  \multicolumn{1}{l|}{} &
  \multicolumn{1}{l|}{} &
  \multicolumn{1}{l|}{} &
  \multicolumn{1}{l|}{BF} &
  \multicolumn{1}{l|}{BF(T)} &
  \multicolumn{1}{l|}{SHG} &
  \multicolumn{1}{l|}{SHG(T)} \\ \cline{4-8} 
\multicolumn{1}{c|}{} &
  \multicolumn{1}{l|}{} &
  \multicolumn{1}{l|}{} &
  \multicolumn{1}{l|}{BF} &
  \cellcolor[HTML]{A1CEF0}40.3 &
  \cellcolor[HTML]{FFD570} 28.7 &
  5.3 &
  \multicolumn{1}{r|}{\cellcolor[HTML]{FF9CA2}3.7} \\ \cline{4-4}
\multicolumn{1}{c|}{} &
  \multicolumn{1}{l|}{} &
  \multicolumn{1}{l|}{\multirow{-3}{*}{Originals}} &
  \multicolumn{1}{l|}{SHG} &
  5.2 &
  \cellcolor[HTML]{FF9CA2}6.0 &
  \cellcolor[HTML]{A1CEF0}24.6 &
  \multicolumn{1}{r|}{\cellcolor[HTML]{FFD570}31.3} \\ \cline{3-8} 
\multicolumn{1}{c|}{} &
  \multicolumn{1}{l|}{} &
  \multicolumn{1}{l|}{} &
  \multicolumn{1}{l|}{} &
  \multicolumn{1}{l|}{CoMIR(BF)} &
  \multicolumn{1}{l|}{CoMIR(BF(T))} &
  \multicolumn{1}{l|}{CoMIR(SHG)} &
  \multicolumn{1}{l|}{CoMIR(SHG(T))} \\ \cline{4-8} 
\multicolumn{1}{c|}{} &
  \multicolumn{1}{l|}{} &
  \multicolumn{1}{l|}{} &
  \multicolumn{1}{l|}{CoMIR(BF)} &
  \cellcolor[HTML]{A1CEF0} 24.6 &
   \cellcolor[HTML]{FFD570} 13.4 & 11.2 &
  \multicolumn{1}{r|}{\cellcolor[HTML]{FF9CA2} 6.0} \\ \cline{4-4}
\multicolumn{1}{c|}{\multirow{-42}{*}{\rotatebox[origin=c]{90}{}}} &
  \multicolumn{1}{l|}{\multirow{-6}{*}{\rotatebox[origin=c]{90}{2DKD}}} &
  \multicolumn{1}{l|}{\multirow{-3}{*}{CoMIR}} &
  \multicolumn{1}{l|}{CoMIR(SHG)} &
   6.7 &
  \cellcolor[HTML]{FF9CA2} 9.7 &
  \cellcolor[HTML]{A1CEF0} 18.7 &
  \multicolumn{1}{r|}{\cellcolor[HTML]{FFD570} 12.7 } \\ \cline{2-8}
\end{tabular}
}
\subcaption{Retrieval results for the full image search.}
\label{tab:FullSize}
%\end{table}
\end{subtable}
%\hfill
% Please add the following required packages to your document preamble:
% \usepackage{multirow}
% \usepackage[table,xcdraw]{xcolor}
% If you use beamer only pass "xcolor=table" option, i.e. \documentclass[xcolor=table]{beamer}
%\begin{table}[t]
%\resizebox{0.49\textwidth}{!}{%
\begin{subtable}[c]{0.49\textwidth}
\centering
\resizebox{\textwidth}{!}{
\begin{tabular}{clllrrrr}
 & & & &
  \multicolumn{4}{c}{\textbf{Query}} \\ \cline{2-8} 
\multicolumn{1}{c|}{} &
  \multicolumn{1}{l|}{} &
  \multicolumn{1}{l|}{} &
  \multicolumn{1}{l|}{} &
  \multicolumn{1}{l|}{BF} &
  \multicolumn{1}{l|}{BF(T)} &
  \multicolumn{1}{l|}{SHG} &
  \multicolumn{1}{l|}{SHG(T)} \\ \cline{4-8} 
\multicolumn{1}{c|}{} &
  \multicolumn{1}{l|}{} &
  \multicolumn{1}{l|}{} &
  \multicolumn{1}{l|}{BF} &
  \cellcolor[HTML]{A1CEF0}97.5 &
  \cellcolor[HTML]{FFD570}97 &
  8.7 &
  \multicolumn{1}{r|}{\cellcolor[HTML]{FF9CA2}8.2} \\ \cline{4-4}
\multicolumn{1}{c|}{} &
  \multicolumn{1}{l|}{} &
  \multicolumn{1}{l|}{\multirow{-3}{*}{Originals}} &
  \multicolumn{1}{l|}{SHG} &
  10.4 &
  \cellcolor[HTML]{FF9CA2}10 &
  \cellcolor[HTML]{A1CEF0}86.3 &
  \multicolumn{1}{r|}{\cellcolor[HTML]{FFD570}82.1} \\ \cline{3-8} 
\multicolumn{1}{c|}{} &
  \multicolumn{1}{l|}{} &
  \multicolumn{1}{l|}{} &
  \multicolumn{1}{l|}{} &
  \multicolumn{1}{l|}{CoMIR(BF)} &
  \multicolumn{1}{l|}{CoMIR(BF(T))} &
  \multicolumn{1}{l|}{CoMIR(SHG)} &
  \multicolumn{1}{l|}{CoMIR(SHG(T))} \\ \cline{4-8} 
\multicolumn{1}{c|}{} &
  \multicolumn{1}{l|}{} &
  \multicolumn{1}{l|}{} &
  \multicolumn{1}{l|}{CoMIR(BF)} &
  \cellcolor[HTML]{A1CEF0}100 &
  \cellcolor[HTML]{FFD570}69.9 &
  17.2 &
  \multicolumn{1}{r|}{{\cellcolor[HTML]{FF9CA2} 18.2}} \\ \cline{4-4}
\multicolumn{1}{c|}{} &
  \multicolumn{1}{l|}{} &
  \multicolumn{1}{l|}{\multirow{-3}{*}{CoMIR}} &
  \multicolumn{1}{l|}{CoMIR(SHG)} &
  18.9 &
  \cellcolor[HTML]{FF9CA2}18.7 &
  \cellcolor[HTML]{A1CEF0}100 &
  \multicolumn{1}{r|}{\cellcolor[HTML]{FFD570}79.1} \\ \cline{3-8} 
\multicolumn{1}{c|}{} &
  \multicolumn{1}{l|}{} &
  \multicolumn{1}{l|}{} &
  \multicolumn{1}{l|}{} &
  \multicolumn{1}{l|}{BF} &
  \multicolumn{1}{l|}{BF(T)} &
  \multicolumn{1}{l|}{Fake BF} &
  \multicolumn{1}{l|}{Fake BF(T)} \\ \cline{4-8} 
\multicolumn{1}{c|}{} &
  \multicolumn{1}{l|}{} &
  \multicolumn{1}{l|}{} &
  \multicolumn{1}{l|}{BF} &
  \cellcolor[HTML]{A1CEF0} 98.9 &
  {\cellcolor[HTML]{FFD570}97.8} &
  11.2 &
  \multicolumn{1}{r|}{\cellcolor[HTML]{FF9CA2}11.2} \\ \cline{4-4}
\multicolumn{1}{c|}{} &
  \multicolumn{1}{l|}{} &
  \multicolumn{1}{l|}{} &
  \multicolumn{1}{l|}{Fake BF} &
  16.4 &
  \cellcolor[HTML]{FF9CA2}17.9 &
  \cellcolor[HTML]{A1CEF0}100 &
  \multicolumn{1}{r|}{\cellcolor[HTML]{FFD570}35.8} \\ \cline{4-8} 
\multicolumn{1}{c|}{} &
  \multicolumn{1}{l|}{} &
  \multicolumn{1}{l|}{} &
  \multicolumn{1}{l|}{} &
  \multicolumn{1}{l|}{SHG} &
  \multicolumn{1}{l|}{SHG (T)} &
  \multicolumn{1}{l|}{Fake SHG} &
  \multicolumn{1}{l|}{Fake SHG(T)} \\ \cline{4-8} 
\multicolumn{1}{c|}{} &
  \multicolumn{1}{l|}{} &
  \multicolumn{1}{l|}{} &
  \multicolumn{1}{l|}{SHG} &
 \cellcolor[HTML]{A1CEF0} 87.7 &
  {\cellcolor[HTML]{FFD570}85.8 }&
  14.2 &
  \multicolumn{1}{r|}{\cellcolor[HTML]{FF9CA2}11.9} \\ \cline{4-4}
\multicolumn{1}{c|}{} &
  \multicolumn{1}{l|}{} &
  \multicolumn{1}{l|}{\multirow{-6}{*}{CycleGAN}} &
  \multicolumn{1}{l|}{Fake SHG} &
  14.2 &
  \cellcolor[HTML]{FF9CA2}15.7 &
  \cellcolor[HTML]{A1CEF0}91.0 &
  \multicolumn{1}{r|}{\cellcolor[HTML]{FFD570}24.6} \\ \cline{3-8} 
\multicolumn{1}{c|}{} &
  \multicolumn{1}{l|}{} &
  \multicolumn{1}{l|}{} &
  \multicolumn{1}{l|}{} &
  \multicolumn{1}{l|}{BF} &
  \multicolumn{1}{l|}{BF(T)} &
  \multicolumn{1}{l|}{Fake BF} &
  \multicolumn{1}{l|}{Fake BF(T)} \\ \cline{4-8} 
\multicolumn{1}{c|}{} &
  \multicolumn{1}{l|}{} &
  \multicolumn{1}{l|}{} &
  \multicolumn{1}{l|}{BF} &
 \cellcolor[HTML]{A1CEF0} 98.9 &
  {\cellcolor[HTML]{FFD570}97.8} &
  14.2 &
  \multicolumn{1}{r|}{\cellcolor[HTML]{FF9CA2}21.6} \\ \cline{4-4}
\multicolumn{1}{c|}{} &
  \multicolumn{1}{l|}{} &
  \multicolumn{1}{l|}{} &
  \multicolumn{1}{l|}{Fake BF} &
  18.7 &
  \cellcolor[HTML]{FF9CA2}23.1 &
  \cellcolor[HTML]{A1CEF0}98.5 &
  \multicolumn{1}{r|}{\cellcolor[HTML]{FFD570}21.6} \\ \cline{4-8} 
\multicolumn{1}{c|}{} &
  \multicolumn{1}{l|}{} &
  \multicolumn{1}{l|}{} &
  \multicolumn{1}{l|}{} &
  \multicolumn{1}{l|}{SHG} &
  \multicolumn{1}{l|}{SHG(T)} &
  \multicolumn{1}{l|}{Fake SHG} &
  \multicolumn{1}{l|}{Fake SHG(T)} \\ \cline{4-8} 
\multicolumn{1}{c|}{} &
  \multicolumn{1}{l|}{} &
  \multicolumn{1}{l|}{} &
  \multicolumn{1}{l|}{SHG} &
 \cellcolor[HTML]{A1CEF0} 87.7 &
  {\cellcolor[HTML]{FFD570}85.8} &
  12.7 &
  \multicolumn{1}{r|}{\cellcolor[HTML]{FF9CA2}9.7} \\ \cline{4-4}
\multicolumn{1}{c|}{} &
  \multicolumn{1}{l|}{\multirow{-18}{*}{\rotatebox[origin=c]{90}{SIFT}}} &
  \multicolumn{1}{l|}{\multirow{-6}{*}{Pix2Pix}} &
  \multicolumn{1}{l|}{Fake SHG} &
  15.7 &
  \cellcolor[HTML]{FF9CA2}15.7 &
  \cellcolor[HTML]{A1CEF0}98.5 &
  \multicolumn{1}{r|}{\cellcolor[HTML]{FFD570}14.2} \\ \cline{2-8}
  \noalign{\vskip\doublerulesep\vskip-\arrayrulewidth} \cline{2-8}
\multicolumn{1}{c|}{} &
  \multicolumn{1}{l|}{} &
  \multicolumn{1}{l|}{} &
  \multicolumn{1}{l|}{} &
  \multicolumn{1}{l|}{BF} &
  \multicolumn{1}{l|}{BF(T)} &
  \multicolumn{1}{l|}{SHG} &
  \multicolumn{1}{l|}{SHG(T)} \\ \cline{4-8} 
\multicolumn{1}{c|}{} &
  \multicolumn{1}{l|}{} &
  \multicolumn{1}{l|}{} &
  \multicolumn{1}{l|}{BF} &
  \cellcolor[HTML]{A1CEF0}95.5 &
  \cellcolor[HTML]{FFD570} 84.8 &
  13.9 &
  \multicolumn{1}{r|}{\cellcolor[HTML]{FF9CA2}13.7} \\ \cline{4-4}
\multicolumn{1}{c|}{} &
  \multicolumn{1}{l|}{} &
  \multicolumn{1}{l|}{\multirow{-3}{*}{Originals}} &
  \multicolumn{1}{l|}{SHG} &
  10.9 &
  \cellcolor[HTML]{FF9CA2}10 &
  \cellcolor[HTML]{A1CEF0}95.8 &
  \multicolumn{1}{r|}{\cellcolor[HTML]{FFD570}88.6} \\ \cline{3-8} 
\multicolumn{1}{c|}{} &
  \multicolumn{1}{l|}{} &
  \multicolumn{1}{l|}{} &
  \multicolumn{1}{l|}{} &
  \multicolumn{1}{l|}{CoMIR(BF)} &
  \multicolumn{1}{l|}{CoMIR(BF(T))} &
  \multicolumn{1}{l|}{CoMIR(SHG)} &
  \multicolumn{1}{l|}{CoMIR(SHG(T))} \\ \cline{4-8} 
\multicolumn{1}{c|}{} &
  \multicolumn{1}{l|}{} &
  \multicolumn{1}{l|}{} &
  \multicolumn{1}{l|}{CoMIR(BF)} &
  \cellcolor[HTML]{A1CEF0}100 &
  \cellcolor[HTML]{FFD570} 76.6 &
  46.3 &
  \multicolumn{1}{r|}{\cellcolor[HTML]{FF9CA2}\textbf{44.8}} \\ \cline{4-4}
\multicolumn{1}{c|}{} &
  \multicolumn{1}{l|}{} &
  \multicolumn{1}{l|}{\multirow{-3}{*}{CoMIR}} &
  \multicolumn{1}{l|}{CoMIR(SHG)} &
  52.2 &
  \cellcolor[HTML]{FF9CA2}\textbf{35.8} &
  \cellcolor[HTML]{A1CEF0}100 &
  \multicolumn{1}{r|}{\cellcolor[HTML]{FFD570}73.6} \\ \cline{3-8} 
\multicolumn{1}{c|}{} &
  \multicolumn{1}{l|}{} &
  \multicolumn{1}{l|}{} &
  \multicolumn{1}{l|}{} &
  \multicolumn{1}{l|}{BF} &
  \multicolumn{1}{l|}{BF(T)} &
  \multicolumn{1}{l|}{Fake BF} &
  \multicolumn{1}{l|}{Fake BF(T)} \\ \cline{4-8} 
\multicolumn{1}{c|}{} &
  \multicolumn{1}{l|}{} &
  \multicolumn{1}{l|}{} &
  \multicolumn{1}{l|}{BF} &
 \cellcolor[HTML]{A1CEF0} 96.3 &
  {\cellcolor[HTML]{FFD570} 88.5} &
  17.9 &
  \multicolumn{1}{r|}{\cellcolor[HTML]{FF9CA2}13.4} \\ \cline{4-4}
\multicolumn{1}{c|}{} &
  \multicolumn{1}{l|}{} &
  \multicolumn{1}{l|}{} &
  \multicolumn{1}{l|}{Fake BF} &
  26.1 &
  \cellcolor[HTML]{FF9CA2}21.6 &
  \cellcolor[HTML]{A1CEF0}98.5 &
  \multicolumn{1}{r|}{\cellcolor[HTML]{FFD570}60.5} \\ \cline{4-8} 
\multicolumn{1}{c|}{} &
  \multicolumn{1}{l|}{} &
  \multicolumn{1}{l|}{} &
  \multicolumn{1}{l|}{} &
  \multicolumn{1}{l|}{SHG} &
  \multicolumn{1}{l|}{SHG(T)} &
  \multicolumn{1}{l|}{Fake SHG} &
  \multicolumn{1}{l|}{Fake SHG(T)} \\ \cline{4-8} 
\multicolumn{1}{c|}{} &
  \multicolumn{1}{l|}{} &
  \multicolumn{1}{l|}{} &
  \multicolumn{1}{l|}{SHG} &
 \cellcolor[HTML]{A1CEF0} 96.7 &
 {\cellcolor[HTML]{FFD570} 91.1} &
  11.9 &
  \multicolumn{1}{r|}{\cellcolor[HTML]{FF9CA2}8.2} \\ \cline{4-4}
\multicolumn{1}{c|}{} &
  \multicolumn{1}{l|}{} &
  \multicolumn{1}{l|}{\multirow{-6}{*}{CycleGAN}} &
  \multicolumn{1}{l|}{Fake SHG} &
  13.4 &
  \cellcolor[HTML]{FF9CA2}13.4 &
  \cellcolor[HTML]{A1CEF0}84.3 &
  \multicolumn{1}{r|}{\cellcolor[HTML]{FFD570}45.5} \\ \cline{3-8} 
\multicolumn{1}{c|}{} &
  \multicolumn{1}{l|}{} &
  \multicolumn{1}{l|}{} &
  \multicolumn{1}{l|}{} &
  \multicolumn{1}{l|}{BF} &
  \multicolumn{1}{l|}{BF(T)} &
  \multicolumn{1}{l|}{Fake BF} &
  \multicolumn{1}{l|}{Fake BF(T)} \\ \cline{4-8} 
\multicolumn{1}{c|}{} &
  \multicolumn{1}{l|}{} &
  \multicolumn{1}{l|}{} &
  \multicolumn{1}{l|}{BF} &
  \cellcolor[HTML]{A1CEF0} 96.3 &
  {\cellcolor[HTML]{FFD570}88.5 }&
  14.9 &
  \multicolumn{1}{r|}{\cellcolor[HTML]{FF9CA2}11.9} \\ \cline{4-4}
\multicolumn{1}{c|}{} &
  \multicolumn{1}{l|}{} &
  \multicolumn{1}{l|}{} &
  \multicolumn{1}{l|}{Fake BF} &
  25.3 &
  \cellcolor[HTML]{FF9CA2}18.7 &
  \cellcolor[HTML]{A1CEF0}97.0 &
  \multicolumn{1}{r|}{\cellcolor[HTML]{FFDD8B}34.3} \\ \cline{4-8} 
\multicolumn{1}{c|}{} &
  \multicolumn{1}{l|}{} &
  \multicolumn{1}{l|}{} &
  \multicolumn{1}{l|}{} &
  \multicolumn{1}{l|}{SHG} &
  \multicolumn{1}{l|}{SHG (T)} &
  \multicolumn{1}{l|}{Fake SHG} &
  \multicolumn{1}{l|}{Fake SHG(T)} \\ \cline{4-8} 
\multicolumn{1}{c|}{} &
  \multicolumn{1}{l|}{} &
  \multicolumn{1}{l|}{} &
  \multicolumn{1}{l|}{SHG} &
  \cellcolor[HTML]{A1CEF0} 96.7 &
  {\cellcolor[HTML]{FFD570}91.1 }&
  16.4 &
  \multicolumn{1}{r|}{\cellcolor[HTML]{FF9CA2}19.4} \\ \cline{4-4}
\multicolumn{1}{c|}{} &
  \multicolumn{1}{l|}{\multirow{-18}{*}{\rotatebox[origin=c]{90}{SURF}}} &
  \multicolumn{1}{l|}{\multirow{-6}{*}{Pix2Pix}} &
  \multicolumn{1}{l|}{Fake SHG} &
  34.3 &
  \cellcolor[HTML]{FF9CA2}34.3 &
  \cellcolor[HTML]{A1CEF0}100 &
  \multicolumn{1}{r|}{\cellcolor[HTML]{FFD570}42.5} \\ \cline{2-8}
  \noalign{\vskip\doublerulesep\vskip-\arrayrulewidth} \cline{2-8}
\multicolumn{1}{c|}{} &
  \multicolumn{1}{l|}{} &
  \multicolumn{1}{l|}{} &
  \multicolumn{1}{l|}{} &
  \multicolumn{1}{l|}{BF} &
  \multicolumn{1}{l|}{BF(T)} &
  \multicolumn{1}{l|}{SHG} &
  \multicolumn{1}{l|}{SHG(T)} \\ \cline{4-8} 
\multicolumn{1}{c|}{} &
  \multicolumn{1}{l|}{} &
  \multicolumn{1}{l|}{} &
  \multicolumn{1}{l|}{BF} &
  \cellcolor[HTML]{A1CEF0}9.0 &
  \cellcolor[HTML]{FFD570} 12.7 &
  11.1 &
  \multicolumn{1}{r|}{\cellcolor[HTML]{FF9CA2}9.0} \\ \cline{4-4}
\multicolumn{1}{c|}{} &
  \multicolumn{1}{l|}{} &
  \multicolumn{1}{l|}{\multirow{-3}{*}{Originals}} &
  \multicolumn{1}{l|}{SHG} &
  11.1 &
  \cellcolor[HTML]{FF9CA2}8.2 &
  \cellcolor[HTML]{A1CEF0}11.9 &
  \multicolumn{1}{r|}{\cellcolor[HTML]{FFD570}10.5} \\ \cline{3-8} 
\multicolumn{1}{c|}{} &
  \multicolumn{1}{l|}{} &
  \multicolumn{1}{l|}{} &
  \multicolumn{1}{l|}{} &
  \multicolumn{1}{l|}{CoMIR(BF)} &
  \multicolumn{1}{l|}{CoMIR(BF(T))} &
  \multicolumn{1}{l|}{CoMIR(SHG)} &
  \multicolumn{1}{l|}{CoMIR(SHG(T))} \\ \cline{4-8} 
\multicolumn{1}{c|}{} &
  \multicolumn{1}{l|}{} &
  \multicolumn{1}{l|}{} &
  \multicolumn{1}{l|}{CoMIR(BF)} &
  \cellcolor[HTML]{A1CEF0} 10.5 &
   \cellcolor[HTML]{FFD570} 12.3 &
   4.1 &
  \multicolumn{1}{r|}{\cellcolor[HTML]{FF9CA2} 7.1} \\ \cline{4-4}
\multicolumn{1}{c|}{\multirow{-42}{*}{\rotatebox[origin=c]{90}{\textbf{Searchable repository}}}} &
  \multicolumn{1}{l|}{\multirow{-6}{*}{\rotatebox[origin=c]{90}{ResNet}}} &
  \multicolumn{1}{l|}{\multirow{-3}{*}{CoMIR}} &
  \multicolumn{1}{l|}{CoMIR(SHG)} &
   8.2 &
  \cellcolor[HTML]{FF9CA2} 10.1 &
  \cellcolor[HTML]{A1CEF0} 14.1 &
  \multicolumn{1}{r|}{\cellcolor[HTML]{FFD570} 13.8} \\ \cline{2-8} \noalign{\vskip\doublerulesep\vskip-\arrayrulewidth} \cline{2-8} % BELOW HERE IS THE 2DKD SECTION
\multicolumn{1}{c|}{} &
  \multicolumn{1}{l|}{} &
  \multicolumn{1}{l|}{} &
  \multicolumn{1}{l|}{} &
  \multicolumn{1}{l|}{BF} &
  \multicolumn{1}{l|}{BF(T)} &
  \multicolumn{1}{l|}{SHG} &
  \multicolumn{1}{l|}{SHG(T)} \\ \cline{4-8} 
\multicolumn{1}{c|}{} &
  \multicolumn{1}{l|}{} &
  \multicolumn{1}{l|}{} &
  \multicolumn{1}{l|}{BF} &
  \cellcolor[HTML]{A1CEF0}26.1 &
  \cellcolor[HTML]{FFD570} 23.8 &
 11.2 &
  \multicolumn{1}{r|}{\cellcolor[HTML]{FF9CA2}6.0} \\ \cline{4-4}
\multicolumn{1}{c|}{} &
  \multicolumn{1}{l|}{} &
  \multicolumn{1}{l|}{\multirow{-3}{*}{Originals}} &
  \multicolumn{1}{l|}{SHG} &
  5.2 &
  \cellcolor[HTML]{FF9CA2}6.0 &
  \cellcolor[HTML]{A1CEF0}24.6 &
  \multicolumn{1}{r|}{\cellcolor[HTML]{FFD570}31.3} \\ \cline{3-8} 
\multicolumn{1}{c|}{} &
  \multicolumn{1}{l|}{} &
  \multicolumn{1}{l|}{} &
  \multicolumn{1}{l|}{} &
  \multicolumn{1}{l|}{CoMIR(BF)} &
  \multicolumn{1}{l|}{CoMIR(BF(T))} &
  \multicolumn{1}{l|}{CoMIR(SHG)} &
  \multicolumn{1}{l|}{CoMIR(SHG(T))} \\ \cline{4-8} 
\multicolumn{1}{c|}{} &
  \multicolumn{1}{l|}{} &
  \multicolumn{1}{l|}{} &
  \multicolumn{1}{l|}{CoMIR(BF)} &
  \cellcolor[HTML]{A1CEF0} 11.2 &
   \cellcolor[HTML]{FFD570} 8.2 & 7.5 &
  \multicolumn{1}{r|}{\cellcolor[HTML]{FF9CA2} 15.7} \\ \cline{4-4}
\multicolumn{1}{c|}{\multirow{-42}{*}{\rotatebox[origin=c]{90}{}}} &
  \multicolumn{1}{l|}{\multirow{-6}{*}{\rotatebox[origin=c]{90}{2DKD}}} &
  \multicolumn{1}{l|}{\multirow{-3}{*}{CoMIR}} &
  \multicolumn{1}{l|}{CoMIR(SHG)} &
   17.9 &
  \cellcolor[HTML]{FF9CA2} 14.9 &
  \cellcolor[HTML]{A1CEF0} 9.0 &
  \multicolumn{1}{r|}{\cellcolor[HTML]{FFD570} 12.0 } \\ \cline{2-8}
\end{tabular}
}
\subcaption{Retrieval results for the image patch search.}
\label{tab:Patch}
\end{subtable}
\caption[Main results of the replacement study]{Main results of the replacement study: Success (in percentage) for top-10 match. BF and SHG represent the two original modalities, \emph{fake SHG} and \emph{fake BF} represent their corresponding I2I representations produced by CycleGAN or pix2pix, and (T) denotes randomly transformed (by rotation and translation) images. Best results for retrieval across modalities and transformations are marked in bold.\label{tab:Results}}
\end{table}

\section*{Results}
\label{sec:results}
\subsection*{Replacement study}
Table~\ref{tab:FullSize} shows the results of the proposed CBIR using full sized query images. It reports the top-10 retrieval success in percentage using the multimodal originals, CoMIRs, and I2I representations produced by CycleGAN and pix2pix as searchable repositories and queries. Experiments were performed using SIFT, SURF and ResNet features to create the vocabulary of the BoW. 
Performance of 2DKD is evaluated on the multimodal original images as well as CoMIRs.
For evaluation of sub-image retrieval in the s-CBIR setup, the same experiments are performed using central patches of $256 \times 256$\,px as query images. Results are shown in Table~\ref{tab:Patch}. Retrieval results of I2I representations in combination with ResNet features and using 2DKD are omitted from the table, as they resulted in fewer retrieval matches than random selection.

\begin{figure}[th!]
  \centering
  \includegraphics[width=0.8\textwidth]{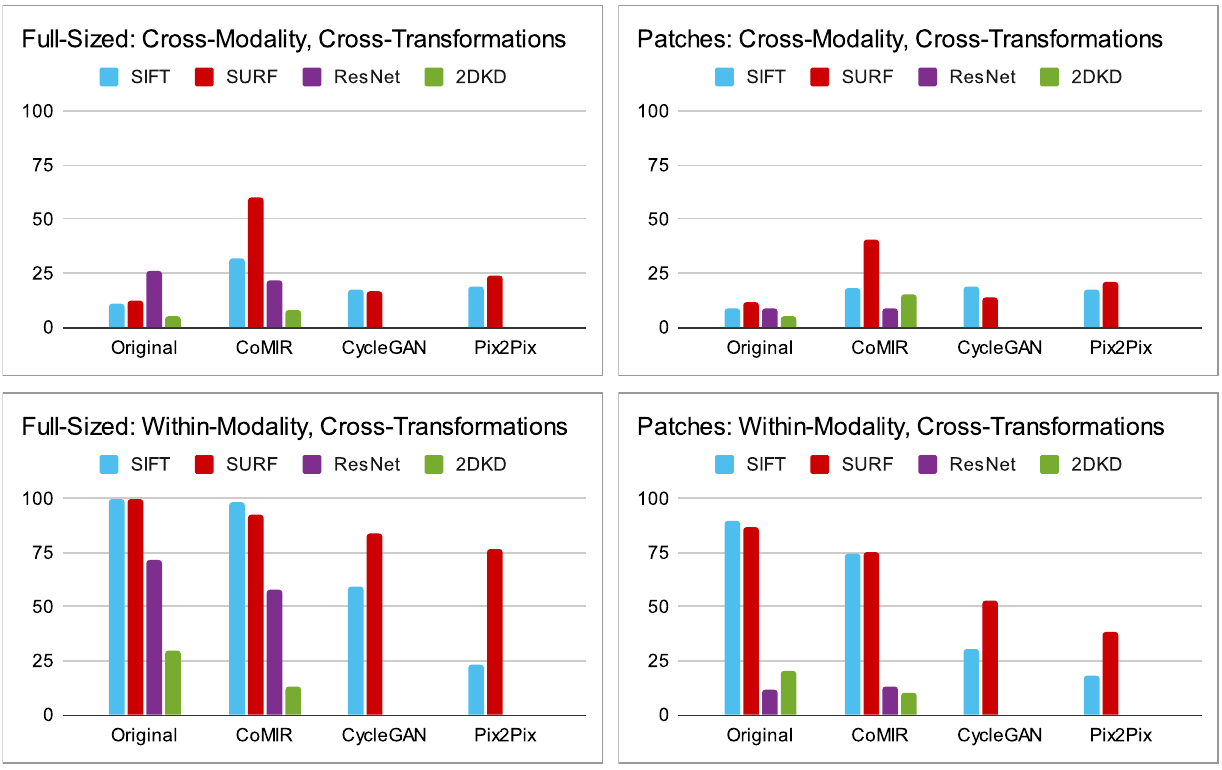}
  \caption[Bar plot of main results of the replacement study]{Top-10 retrieval results without re-ranking, for full-sized image queries (left column) and patches (right column), averaged over retrieval directions (BF query within SHG and SHG query within BF, or their respective representations) for different combinations of images or their learned representations and feature extractors. Cross-modality (top row) and within-modality (bottom row).}
  \label{fig:barplot}
\end{figure}

In Table~\ref{tab:Results}, \textbf{Red cells} present the results of cross-modality retrieval of a transformed (rotated and translated) image in one modality among a set of untransformed images \textit{of the other modality}, which is the main use-case targeted by this study (denoted within-modality, cross-transformations subsequently in Fig. \ref{fig:barplot} and Table \ref{tab:ReRankFull}). 
\textbf{Orange cells} present the results of retrieving a transformed query image within the \textit{same modality} of untransformed images. This gives insight into the invariance of the feature extraction, or the equivariance of the representations under these transformations (denoted within-modality, cross-transformations subsequently in Fig. \ref{fig:barplot} and Table \ref{tab:ReRankFull}).
\textbf{Blue cells} present the results of searching an untransformed query image within the \textit{same modality} of untransformed images, which validates the CBIR setup. A near-perfect retrieval accuracy indicates that the features extracted to create the BoW and its vocabulary size are reasonable.
\textbf{White cells} present the results of retrieving an untransformed query image among images of \textit{the other modality}, which shows how well the learned representations are bridging the semantic gap between the modalities.

Figure~\ref{fig:barplot} summarizes the results for full-sized and patch queries, with numbers from Tables~\ref{tab:FullSize} and \ref{tab:Patch} averaged over retrieval directions (BF query within SHG and SHG query within BF).

\textbf{Re-Ranking}: Re-ranking is evaluated only as a part of the proposed pipeline: with using SURF on CoMIRs to create a BoW. 
In Table~\ref{tab:ReRankFull}, we see that re-ranking among the top-15 retrieval matches can boost the top-10 retrieval performance for full-sized images, increasing retrieval accuracy to 62.0\% (BF as query) and 66.4\% (SHG as query); or 75.4\% and 83.6\% respectively for re-ranking among the top-30 matches.
Table~\ref{tab:ReRankPatch} shows that re-ranking among the top-15 retrieval matches can boost the top-10 retrieval performance also for patch queries, increasing retrieval accuracy to 41.8\% (BF as query) and 53.7\% (SHG as query); or to 51.5\% and 63.4\% respectively for re-ranking among the top-30 matches.

\begin{table}[b]
\begin{subtable}[c]{0.49\textwidth}
\resizebox{\textwidth}{!}{
\begin{tabular}{
>{\columncolor[HTML]{ECE8E8}}c lrrrr}
\multicolumn{1}{l}{\cellcolor[HTML]{ECE8E8}\textbf{Full Sized}} & \multicolumn{5}{c}{\cellcolor[HTML]{ECE8E8}\textbf{Cross-Modality, Cross-Transformations}} \\ \cline{2-6} 
\multicolumn{1}{l|}{\cellcolor[HTML]{ECE8E8}} & \multicolumn{5}{c}{query: CoMIR(BF(T))} \\ \cline{2-6} 
\multicolumn{1}{l|}{\cellcolor[HTML]{ECE8E8}} & \multicolumn{1}{l|}{} & \multicolumn{1}{l}{top-1} & \multicolumn{1}{l}{top-5} & \multicolumn{1}{l}{top-10} & \multicolumn{1}{l}{top-15} \\ \cline{2-6} 
\multicolumn{1}{c|}{\cellcolor[HTML]{ECE8E8}} & \multicolumn{1}{l|}{No Re-Ranking} & 26.9 & 51.5 & 59.0 & 63.4 \\
\multicolumn{1}{c|}{\cellcolor[HTML]{ECE8E8}} & \multicolumn{1}{l|}{Re-Ranking 15} & 32.8 & 55.2 & 62.0 & 63.4 \\
\multicolumn{1}{c|}{\multirow{-3}{*}{\cellcolor[HTML]{ECE8E8}\textbf{CoMIR(SHG)}}} & \multicolumn{1}{l|}{Re-Ranking 30} & 34.3 & 59.7 & 75.4 & 75.4 \\ \cline{2-6} 
\multicolumn{1}{l|}{\cellcolor[HTML]{ECE8E8}} & \multicolumn{5}{c}{query: CoMIR(SHG(T))} \\ \cline{2-6} 
\multicolumn{1}{c|}{\cellcolor[HTML]{ECE8E8}} & \multicolumn{1}{l|}{No Re-Ranking} & 21.6 & 45.5 & 61.2 & 69.4 \\
\multicolumn{1}{c|}{\cellcolor[HTML]{ECE8E8}} & \multicolumn{1}{l|}{Re-Ranking 15} & 42.5 & 57.5 & 66.4 & 69.4 \\
\multicolumn{1}{c|}{\multirow{-3}{*}{\cellcolor[HTML]{ECE8E8}\textbf{CoMIR(BF)}}} & \multicolumn{1}{l|}{Re-Ranking 30} & 45.5 & 64.9 & 83.6 & 83.6
\end{tabular}}
\subcaption{Full size queries.}
\label{tab:ReRankFull}
\end{subtable}
\begin{subtable}[c]{0.49\textwidth}
\resizebox{\textwidth}{!}{%
\begin{tabular}{
>{\columncolor[HTML]{ECE8E8}}c lrrrr}
\multicolumn{1}{l}{\cellcolor[HTML]{ECE8E8}\textbf{Patches}} & \multicolumn{5}{c}{\cellcolor[HTML]{ECE8E8}\textbf{Cross-Modality, Cross-Transformations}} \\ \cline{2-6} 
\multicolumn{1}{l|}{\cellcolor[HTML]{ECE8E8}} & \multicolumn{5}{c}{query: CoMIR(BF(T))} \\ \cline{2-6} 
\multicolumn{1}{l|}{\cellcolor[HTML]{ECE8E8}} & \multicolumn{1}{l|}{} & \multicolumn{1}{l}{top-1} & \multicolumn{1}{l}{top-5} & \multicolumn{1}{l}{top-10} & \multicolumn{1}{l}{top-15} \\ \cline{2-6} 
\multicolumn{1}{c|}{\cellcolor[HTML]{ECE8E8}} & \multicolumn{1}{l|}{No Re-Ranking} & 11.2 & 25.4 & 35.8 & 42.5 \\
\multicolumn{1}{c|}{\cellcolor[HTML]{ECE8E8}} & \multicolumn{1}{l|}{Re-Ranking 15} & 23.1 & 37.3 & 41.8 & 42.5 \\
\multicolumn{1}{c|}{\multirow{-3}{*}{\cellcolor[HTML]{ECE8E8}\textbf{CoMIR(SHG)}}} & \multicolumn{1}{l|}{Re-Ranking 30} & 27.6 & 47.0 & 51.5 & 55.9 \\ \cline{2-6} 
\multicolumn{1}{l|}{\cellcolor[HTML]{ECE8E8}} & \multicolumn{5}{c}{query: CoMIR(SHG(T))} \\ \cline{2-6} 
\multicolumn{1}{c|}{\cellcolor[HTML]{ECE8E8}} & \multicolumn{1}{l|}{No Re-Ranking} & 10.4 & 32.8 & 44.8 & 53.7 \\
\multicolumn{1}{c|}{\cellcolor[HTML]{ECE8E8}} & \multicolumn{1}{l|}{Re-Ranking 15} & 29.1 & 46.3 & 53.7 & 53.7 \\
\multicolumn{1}{c|}{\multirow{-3}{*}{\cellcolor[HTML]{ECE8E8}\textbf{CoMIR(BF)}}} & \multicolumn{1}{l|}{Re-Ranking 30} & 29.1 & 53.7 & 63.4 & 67.2
\end{tabular}}
\subcaption{Patch queries.}
\label{tab:ReRankPatch}
\end{subtable}
\caption[Performance comparison with and without re-ranking]{Performance gain for full size (left) and patch queries (right) due to re-ranking among the top-15 and top-30 matches of the main pipeline using SURF features on CoMIRs.}
\end{table}

\subsection*{Comparison with state-of-the-art}

Among the latest state-of-the-art methods in CMIR, few are applicable to instance-level retrieval between modalities as different as the ones used in our evaluation dataset. IMTDF relies not only on ResNet features, but also on a number of handcrafted features that perform well for within-modality retrieval on medical images. As seen in Table~\ref{tab:sta} however, this combination of features does not perform well for cross-modality retrieval of BF and SHG images. 
TC-Net on the other hand has been previously evaluated on modalities very different from BF and SHG, but is generally not domain specific. It is based on a similar training mechanism as the representation learning stage in our proposed method and outperforms IMTDF on retrieval across BF and SHG (see Table~\ref{tab:sta}). However, our proposed method outperforms both TC-Net and IMTDF, even without re-ranking.

%while there is a number of newer and better performing methods available for sketch&img retrieval, this is the latest one we found that was directly applicable to our evaluation use case. 

\begin{table}[b]
\begin{subtable}[c]{0.49\textwidth}
\resizebox{\textwidth}{!}{
\begin{tabular}{lclcccc}
 &  &  & \textbf{} & \multicolumn{2}{c}{\textbf{{\scriptsize Query}}} & \multicolumn{1}{l}{} \\ \hhline{~|~|~|-|-|-|-|}%\cline{4-7} 
 & \multicolumn{1}{l}{\cellcolor[HTML]{FFFFFF}} & \multicolumn{1}{l|}{\cellcolor[HTML]{FFFFFF}} & \multicolumn{1}{l}{\cellcolor[HTML]{FFFFFF}BF} & \multicolumn{1}{l}{\cellcolor[HTML]{FFFFFF}BF(T)} & \multicolumn{1}{l}{\cellcolor[HTML]{FFFFFF}SHG} & \multicolumn{1}{l|}{\cellcolor[HTML]{FFFFFF}SHG(T)} \\ \hhline{~|-|-|-|-|-|-|} %\cline{2-7} 
\multicolumn{1}{l|}{} & \multicolumn{1}{c|}{\cellcolor[HTML]{ECE8E8}} & \multicolumn{1}{l|}{\cellcolor[HTML]{FFFFFF}BF} & - & - & \multicolumn{1}{r}{89.6} & \multicolumn{1}{r|}{83.6} \\
\multicolumn{1}{l|}{} & \multicolumn{1}{c|}{\multirow{-2}{*}{\cellcolor[HTML]{ECE8E8}\textbf{Proposed}}} & \multicolumn{1}{l|}{\cellcolor[HTML]{FFFFFF}SHG} & \multicolumn{1}{r}{86.6} & \multicolumn{1}{r}{75.4} & - & \multicolumn{1}{c|}{-} \\ \hhline{~|-|-|-|-|-|-|} %\cline{2-7} 
\multicolumn{1}{l|}{} & \multicolumn{1}{c|}{\cellcolor[HTML]{ECE8E8}} & \multicolumn{1}{l|}{\cellcolor[HTML]{FFFFFF}BF} & - & - & \multicolumn{1}{r}{47.8} & \multicolumn{1}{r|}{43.4} \\
\multicolumn{1}{l|}{} & \multicolumn{1}{c|}{\multirow{-2}{*}{\cellcolor[HTML]{ECE8E8}\textbf{TC-Net}}} & \multicolumn{1}{l|}{\cellcolor[HTML]{FFFFFF}SHG} & \multicolumn{1}{r}{35.8} & \multicolumn{1}{r}{37.3} & - & \multicolumn{1}{c|}{-} \\ \hhline{~|-|-|-|-|-|-|} %cline{2-7} 
\multicolumn{1}{l|}{} & \multicolumn{1}{c|}{\cellcolor[HTML]{ECE8E8}} & \multicolumn{1}{l|}{\cellcolor[HTML]{FFFFFF}BF} & - & - & \multicolumn{1}{r}{9.7} & \multicolumn{1}{r|}{11.9} \\
\multicolumn{1}{l|}{\multirow{-6}{*}{{\rotatebox[origin=c]{90}{\textbf{{\scriptsize Searchable Repository}}}}}} & \multicolumn{1}{c|}{\multirow{-2}{*}{\cellcolor[HTML]{ECE8E8}\textbf{IMTDF}}} & \multicolumn{1}{l|}{\cellcolor[HTML]{FFFFFF}SHG} & \multicolumn{1}{r}{12.7} & \multicolumn{1}{r}{11.2} & - & \multicolumn{1}{c|}{-} \\ \hhline{~|-|-|-|-|-|-|} %\cline{2-7} 
\end{tabular}}
\subcaption{Full size queries.}
\label{tab:FullCompete}
\end{subtable}
\begin{subtable}[c]{0.49\textwidth}
\resizebox{\textwidth}{!}{%
\begin{tabular}{lclcccc}
 &  &  & \textbf{} & \multicolumn{2}{c}{\textbf{{\scriptsize Query}}} & \multicolumn{1}{l}{} \\ \hhline{~|~|~|-|-|-|-|} %\cline{4-7} 
 & \multicolumn{1}{l}{\cellcolor[HTML]{FFFFFF}} & \multicolumn{1}{l|}{\cellcolor[HTML]{FFFFFF}} & \multicolumn{1}{l}{\cellcolor[HTML]{FFFFFF}BF} & \multicolumn{1}{l}{\cellcolor[HTML]{FFFFFF}BF(T)} & \multicolumn{1}{l}{\cellcolor[HTML]{FFFFFF}SHG} & \multicolumn{1}{l|}{\cellcolor[HTML]{FFFFFF}SHG(T)} \\ \hhline{~|-|-|-|-|-|-|} %\cline{2-7} 
\multicolumn{1}{l|}{} & \multicolumn{1}{c|}{\cellcolor[HTML]{ECE8E8}} & \multicolumn{1}{l|}{\cellcolor[HTML]{FFFFFF}BF} & - & - & \multicolumn{1}{r}{66.4} & \multicolumn{1}{r|}{63.4} \\
\multicolumn{1}{l|}{} & \multicolumn{1}{c|}{\multirow{-2}{*}{\cellcolor[HTML]{ECE8E8}\textbf{Proposed}}} & \multicolumn{1}{l|}{\cellcolor[HTML]{FFFFFF}SHG} & \multicolumn{1}{r}{56.7} & \multicolumn{1}{r}{51.5} & - & \multicolumn{1}{c|}{-} \\ \hhline{~|-|-|-|-|-|-|} %\cline{2-7} 
\multicolumn{1}{l|}{} & \multicolumn{1}{c|}{\cellcolor[HTML]{ECE8E8}} & \multicolumn{1}{l|}{\cellcolor[HTML]{FFFFFF}BF} & - & - & \multicolumn{1}{r}{22.4} & \multicolumn{1}{r|}{20.1} \\
\multicolumn{1}{l|}{} & \multicolumn{1}{c|}{\multirow{-2}{*}{\cellcolor[HTML]{ECE8E8}\textbf{TC-Net}}} & \multicolumn{1}{l|}{\cellcolor[HTML]{FFFFFF}SHG} & \multicolumn{1}{r}{12.7} & \multicolumn{1}{r}{16.4} & - & \multicolumn{1}{c|}{-} \\ \hhline{~|-|-|-|-|-|-|} %\cline{2-7} 
\multicolumn{1}{l|}{} & \multicolumn{1}{c|}{\cellcolor[HTML]{ECE8E8}} & \multicolumn{1}{l|}{\cellcolor[HTML]{FFFFFF}BF} & - & - & \multicolumn{1}{r}{7.5} & \multicolumn{1}{r|}{7.5} \\
\multicolumn{1}{l|}{\multirow{-6}{*}{{\rotatebox[origin=c]{90}{\textbf{{\scriptsize Searchable Repository}}}}}} & \multicolumn{1}{c|}{\multirow{-2}{*}{\cellcolor[HTML]{ECE8E8}\textbf{IMTDF}}} & \multicolumn{1}{l|}{\cellcolor[HTML]{FFFFFF}SHG} & \multicolumn{1}{r}{7.5} & \multicolumn{1}{r}{8.2} & - & \multicolumn{1}{c|}{-} \\ \hhline{~|-|-|-|-|-|-|} %\cline{2-7} 
\end{tabular}}
\subcaption{Patch queries.}
\label{tab:PatchCompete}
\end{subtable}
\caption[Results of comparison to state-of-the-art methods]{Top-10 retrieval success (in percentage) of two state-of-the-art RIS methods on (sub-)image retrieval across BF\&SHG modalities. Here reported retrieval success of our proposed pipeline includes top-30 reranking.}
\label{tab:sta}
\end{table}

\section*{Discussion} 

Our study demonstrates that out of all tested settings and methods, our proposed pipeline is the best-performing one for the task at hand, yielding a 61.2\% top-10 success rate retrieving BF queries in a set of SHG images and 59.0\% retrieving SHG queries within the set of BF images. With re-ranking the first 30 matches, these results are further improved to 75.4\% and 83.6\% respectively. The strength of combining learning based CoMIRs with classic feature extractors such as SURF, merges the potential of CNNs to produce equivariant representations which can bridge between different modalities, and robust, sparse feature extractors that are rotationally and translationally invariant, and due to their speed qualify for creating (s-)CBIR systems for large datasets. Moreover, CoMIR and SURF are modality-independent, not incorporating any data-specific information, which makes the pipeline more generally applicable.

\begin{figure}[ht!]
  \centering
  \begin{subfigure}[t]{0.65\textwidth}
  \includegraphics[width=\textwidth]{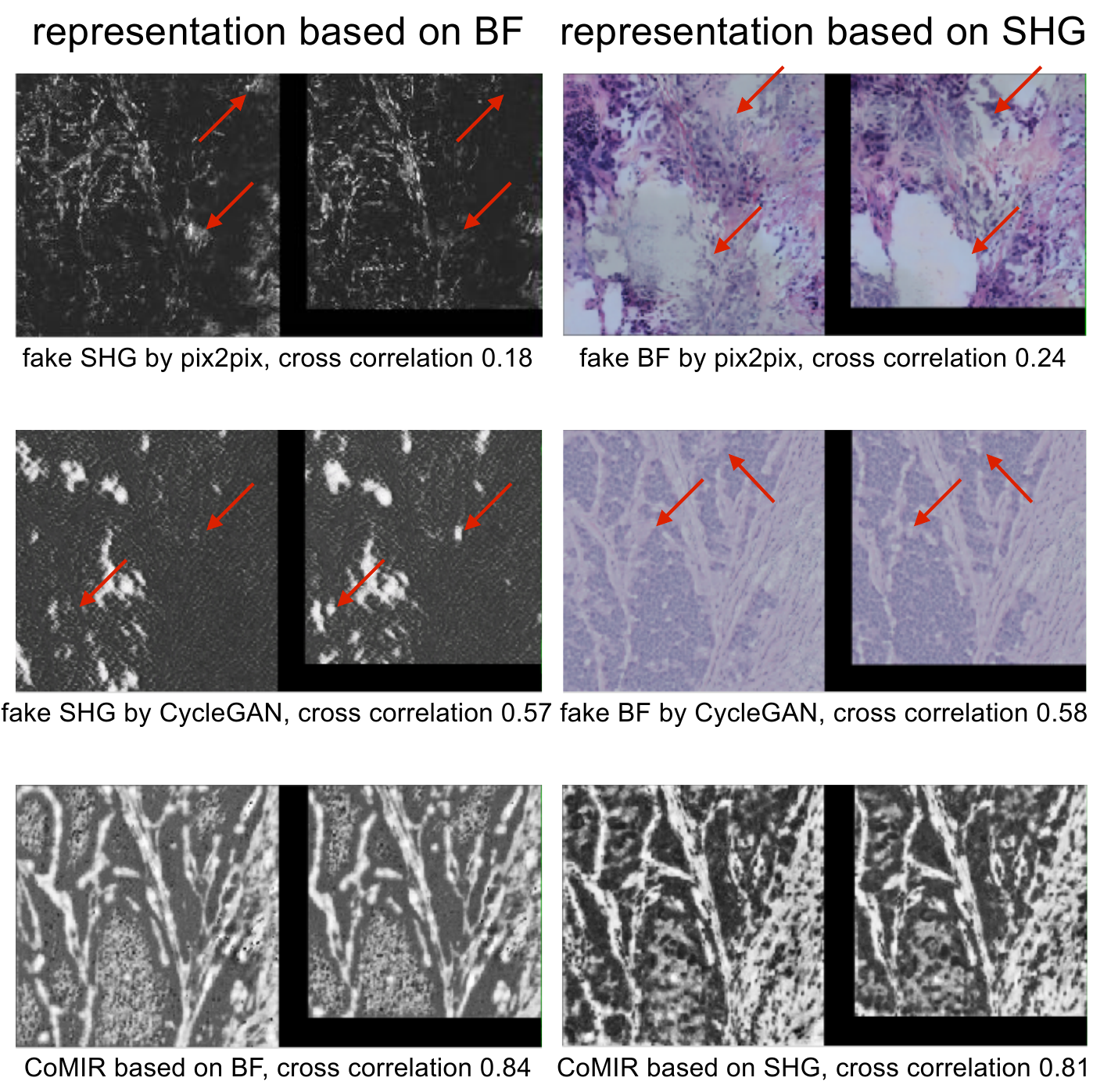}
  \caption{Representations produced by I2I and CoMIR. The image pairs to the left show the learned representations for one BF image in the test set;  first column corresponds to the untransformed, and the second to the transformed image, aligned back for comparison. 
  %representation of the untransformed image, second corresponds to the transformed aligned image. 
  The pairs on the right show the representations of the SHG image;  third column corresponds to the untransformed, and the fourth to the transformed aligned image. Below the pairs, the 2D correlation coefficient for their overlapping area is given. Red arrows guide the reader to locations in the images in which structures clearly differ.}
  \label{fig:GAN}
\end{subfigure}
~
\begin{subfigure}[t]{0.32\textwidth}
  \includegraphics[width=\textwidth]{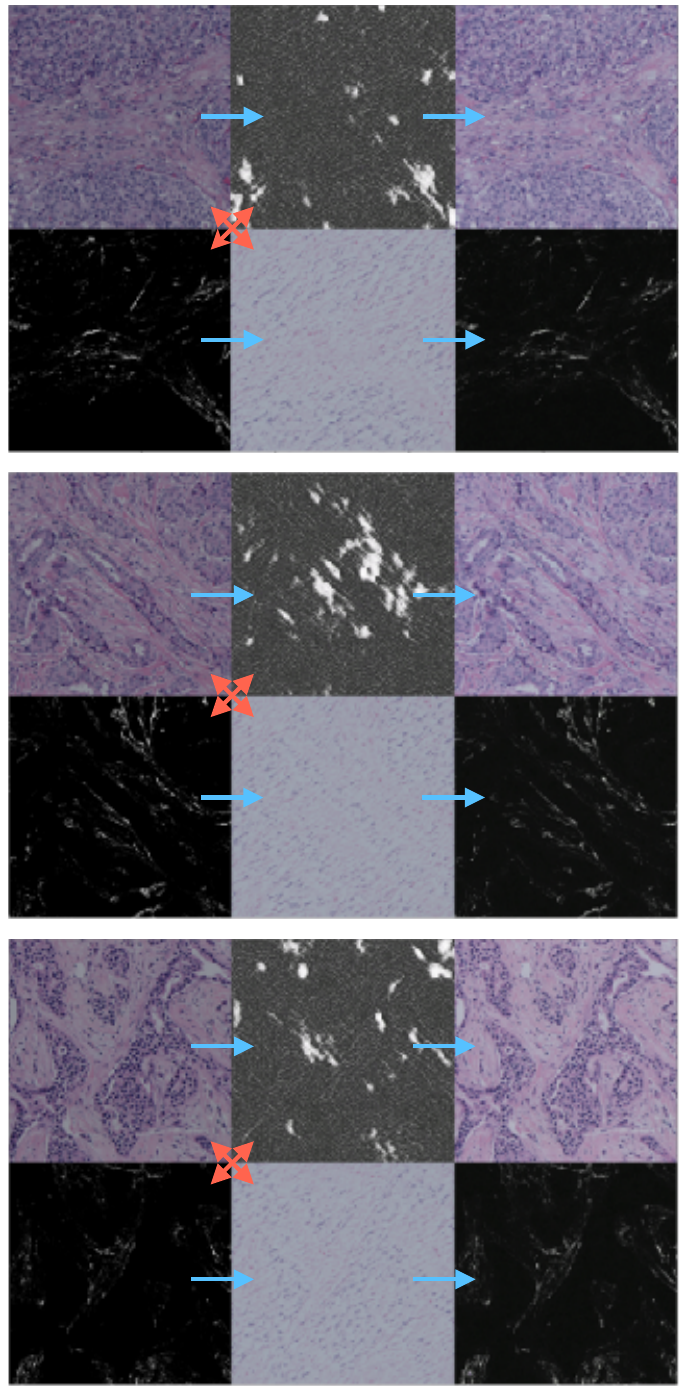}
  \caption{The first column shows original BF \& SHG images, the second shows CycleGAN's representations in the respective other modality given those originals (indicated by the blue arrow). The third column shows the reconstructed originals. Red arrows indicate image pairs which should be similar.}
  \label{fig:CycleGAN}
  \end{subfigure}
  \caption[Visual examples of the test set demonstrating the shortcomings of the I2I approaches]{Visual examples of the test set demonstrating the shortcomings of the I2I approaches to bridge the semantic gap between the modalities. (a) shows that translational and rotational equivariance is not preserved for I2I generated images, (b) shows that the fake BF images (even rows, middle) do not preserve the structure and appearance of the corresponding real BF images, but appear similar, independent of the content of the SHG images they are generated from.}
\end{figure}

As seen in Table~\ref{tab:FullSize}, performing within-modality full image retrieval on the original images with transformed queries has a 100\% or close to 100\% success rate when using the SURF or SIFT feature extractors, while using ResNet as a feature extractor results in a significant drop in retrieval success. This can be attributed to the lack of rotational invariance of ResNet as a feature extractor. While this high within-modality retrieval success is retained when using CoMIR embeddings (i.e. retrieving the CoMIR of a transformed image in modality \textit{A} within the set of CoMIRs of untransformed images of modality \textit{A} and vice versa), it drops significantly with the use of I2I approaches (i.e. retrieving the fake GAN image of a transformed image in modality \textit{A} within the set of CoMIRs of untransformed images of modality \textit{A} and vice versa) even when using SURF or SIFT features.  Since SIFT and SURF are rotationally invariant by design, we argue that the reduced performance when using them in combination with I2I approaches is due to the GAN-generated images not preserving translational, and in particular rotational, equivariances in their representations. 
The reason behind this shortcoming of the GAN-generated images is the absence of network architecture related enforcement (as there is with, e.g., steerable CNNs or group convolutions) in place for pix2pix or CycleGAN to relate rotated versions of the input with each other. As long as their generated fake images belong to the distribution of the target modality, the discriminator will accept them as reasonable output. Figure~\ref{fig:GAN} shows an example from the test set illustrating this effect. As seen by the cross-correlation of their (aligned) overlap, the fake representations of the untransformed and transformed images can differ significantly.

On the other hand, attempting cross-modality retrieval directly on the original images fails regardless of the choice of feature extractor, thus highlighting the need for bringing the two modalities closer together. While bridging the gap between SHG and BF modalities through CoMIRs 
%enables us to strongly improve the
delivers strongly improved
cross-modality retrieval success when using SIFT or SURF, the I2I approaches are less advantageous. We notice that CycleGAN suffers from so-called mode-collapse. In Figure~\ref{fig:CycleGAN}, three examples are shown for which the fake BF modalities (middle image in row 2,4 and 6) are extremely similar, independent of the input images. The structures in the original BF images are not preserved, instead texture was generated that is accepted by the discriminator as a reasonable BF tissue. While these fake BF images successfully encode the information required to reconstruct the SHG images, fulfilling the cycle consistency (third column in Figure 6), they failed to produce a representation similar to the real target BF image.

The rotationally equivariant CoMIRs together with invariant feature extractors like SIFT and SURF can handle the displacements between the images, and the representations suffice to bridge between the modalities of SHG and BF. The best results for cross-modality retrieval of transformed full-sized images are obtained by our porposed method, using CoMIRs to learn representations for both input modalities, in combination with SURF to extract features for the BoW. 

Similar behaviour to full sized image retrieval can be seen also in Table~\ref{tab:Patch}, when querying patches. However for s-CBIR, ResNet as a dense feature extractor is not able to compete with SIFT and SURF.  It extracts the same number of features regardless of the input image size. Hence, the resulting feature descriptor is of the same dimension for both the full sized images in the searchable repository, as well as for the query image. The average pooling layer of the network blurs out the features which result from the common region of the full sized image and the patch. This highlights the advantage of sparse feature extractors like SIFT and SURF.

Comparing the use of BoW models to a recently introduced RIS toolkit 2DKD (originally developed for within modality search), shows that the performance of 2DKD for cross-modal retrieval is higher using CoMIRs than using the original images, but the results are still significantly worse than the BoW based approaches. Even the within-modality retrieval performance of 2DKD is low. This is likely due to 2DKD extracting descriptors based on Krawtchouk polynomials, which can be seen as shape descriptors. The BF images used in this study are dense and their content corresponds rather to texture than shape, whereas SHG images are sparse and lack concrete shapes.

Furthermore, the proposed method outperforms the recent CBIR successfully used in medical image retrieval IMTDF, and the recent cross-modal image retrieval method TC-Net (Table~\ref{tab:sta}). Similarly to the representation learning of CoMIRs used in our method, TC-NET uses a contrastive loss (triplet loss), but learns the feature embeddings used for retrieval directly. However, it uses a siamese network, i.e. all weights are shared for the networks streams processing the different modalities. We suspect that the weight sharing is the reason behind the lower retrieval success of TC-Net, as it can degrade performance when the modalities are very different in structure, as is the case for BF and SHG.

%The comparison to more recent method medical image retrieval (in Table~\ref{tab:sta}) shows that our proposed pipeline enables significantly more accurate retrieval across modalities then the latest available alternatives. We suspect that the reasons behind the lower retrieval success of TC-Net despite its similar structure is the utilization of weight-sharing in all three branches of the network. Weight sharing can degrade performance when the modalities are very different structure-wise, as is the case for BF and SHG. %additional disadvantage: the use a set of losses and weight them, which results in a large number of hyperparameters that may be domain- or even individual dataset-dependent. 

\section*{Conclusion}
\label{sec:concl}
We present a novel, data-independent approach for 
the challenging task of instance-level reverse image search across modalities without labels, and evaluate it on BF and SHG microscopy images used in histopathology. We combine the power of deep learning to generate representations for images of different modalities, with robust classical methods of feature extraction to create a BoW. Finally, we add re-ranking to further boost the retrieval success. 
Our proposed method outperforms two most recent approaches applicable to instance-level retrieval across BF and SHG modalities.
Through a replacement study we confirm its efficacy and superiority over other design choices. 
We observe that using shape descriptors relying on Krawtchouk moments is inferior to BoW models for retrieval of BF and SHG images, and that in order to apply representation learning or I2I to bridge between modalities, it is essential that the learned representations are equivariant under the transformations between corresponding images. Furthermore, we show that it is crucial to use translation- and rotation-invariant feature extractors such as SIFT and SURF.
Future work includes testing the pipeline on other modalities, and developing an improved feature selection and matching procedure tailored specifically for CoMIR type representations. In addition, the interplay between the raw modalities and their CoMIRs can be further explored, to assess if a fused utilization can yield further improvements.

\section*{Acknowledgements}
E.B. and E.W. are partly funded by the CIM, Uppsala University. Vinnova, Sweden’s Innovation Agency projects 2017-02447 partly funds EW, JL, NS and 2020-03611 JL and NS. All figures in this study were created by the authors.

\subsection*{Author contributions statement}

E.B. and E.W. wrote the code, conducted the experiments and analyzed the results.  All authors were involved in writing and reviewing of the manuscript. 

\subsection*{Data availability}
The dataset used in the study is freely accessible in the zenodo repository, \url{https://zenodo.org/record/3874362}.

\subsection*{Competing interests}
The authors declare no competing interests. 

%\subsection*{Additional information}
%To include, in this order: \textbf{Accession codes} (where applicable); \textbf{Competing interests} (mandatory statement). 

%The corresponding author is responsible for submitting a \href{http://www.nature.com/srep/policies/index.html#competing}{competing interests statement} on behalf of all authors of the paper. This statement must be included in the submitted article file.

\bibliographystyle{plain}
\bibliography{main}

\begin{thebibliography}{10}

\bibitem{babenko2015}
A.~Babenko and V.~Lempitsky.
\newblock Aggregating local deep features for image retireval.
\newblock In {\em Intl. Conf. on Computer Vision (ICCV)}, 2015.

\bibitem{BAI2020102835}
Cong Bai, Jian Chen, Qing Ma, Pengyi Hao, and Shengyong Chen.
\newblock Cross-domain representation learning by domain-migration generative
  adversarial network for sketch based image retrieval.
\newblock {\em Journal of Visual Communication and Image Representation},
  71:102835, 2020.

\bibitem{nephro}
Laura Barisoni, Kyle~J. Lafata, Stephen~M. Hewitt, Anant Madabhushi, and
  Ulysses G.~J. Balis.
\newblock Digital pathology and computational image analysis in
  nephropathology.
\newblock {\em Nat Rev Nephrol}, 16:669--685, 2020.

\bibitem{SURF}
H.~Bay, A.~Ess, T.~Tuytelaars, and L.~Van~Gool.
\newblock {SURF}: Speeded up robust features.
\newblock {\em Computer Vision and Image Understanding (CVIU)},
  110(3):346–359, 2008.

\bibitem{satelit2019}
Vijayakumar Bhandi and K.~A. Sumithra~Devi.
\newblock Image retrieval by fusion of features from pre-trained deep
  convolution neural networks.
\newblock In {\em Intl. Conf. on Advanced Technologies in Intelligent Control,
  Environment, Computing Communication Engineering (ICATIECE)}, pages 35--40,
  2019.

\bibitem{BUI201727}
T.~Bui, L.~Ribeiro, M.~Ponti, and J.~Collomosse.
\newblock Compact descriptors for sketch-based image retrieval using a triplet
  loss convolutional neural network.
\newblock {\em Computer Vision and Image Understanding}, 164:27--37, 2017.

\bibitem{10.1007/978-3-642-02976-9_17}
Juan~C. Caicedo, Angel Cruz, and Fabio~A. Gonzalez.
\newblock Histopathology image classification using bag of features and kernel
  functions.
\newblock In {\em Artificial Intelligence in Medicine}, pages 126--135, Berlin,
  Heidelberg, 2009. Springer Berlin Heidelberg.

\bibitem{10.1007/978-3-030-58565-5_43}
Bingyi Cao, Andr{\'e} Araujo, and Jack Sim.
\newblock Unifying deep local and global features for image search.
\newblock In Andrea Vedaldi, Horst Bischof, Thomas Brox, and Jan-Michael Frahm,
  editors, {\em Computer Vision -- ECCV 2020}, pages 726--743, Cham, 2020.
  Springer Intl. Publishing.

\bibitem{CHEN2020105630}
Pingjun Chen, Xiaoshuang Shi, Yun Liang, Yuan Li, Lin Yang, and Paul~D. Gader.
\newblock Interactive thyroid whole slide image diagnostic system using deep
  representation.
\newblock {\em Computer Methods and Programs in Biomedicine}, 195:105630, 2020.

\bibitem{2KDK}
Julian~S DeVille, Daisuke Kihara, and Atilla Sit.
\newblock {2DKD}: a toolkit for content-based local image search.
\newblock {\em Source Code Biol Med.}, 2020.

\bibitem{kevin_eliceiri_2020_3874362}
Kevin Eliceiri, Bin Li, and Adib Keikhosravi.
\newblock Multimodal biomedical dataset for evaluating registration methods
  (patches from {TMA} cores).
\newblock \emph{zenodo} \url{https://zenodo.org/record/3874362}, June 2020.

\bibitem{FANG2021101981}
Jiansheng Fang, Huazhu Fu, and Jiang Liu.
\newblock Deep triplet hashing network for case-based medical image retrieval.
\newblock {\em Medical Image Analysis}, 69:101981, 2021.

\bibitem{7780459}
Kaiming He, Xiangyu Zhang, Shaoqing Ren, and Jian Sun.
\newblock Deep residual learning for image recognition.
\newblock In {\em 2016 IEEE Conference on Computer Vision and Pattern
  Recognition (CVPR)}, pages 770--778, 2016.

\bibitem{smily}
N.~Hedge, J.D. Hipp, and Y.~Liu.
\newblock Similar image search for histology: {SMILY}.
\newblock {\em npj Digit. Med.}, 2(56), 2019.

\bibitem{Hristu:21}
Radu Hristu, Stefan~G. Stanciu, Adrian Dumitru, Bogdan Paun, Iustin Floroiu,
  Mariana Costache, and George~A. Stanciu.
\newblock Influence of hematoxylin and eosin staining on the quantitative
  analysis of second harmonic generation imaging of fixed tissue sections.
\newblock {\em Biomed. Opt. Express}, 12(9):5829--5843, Sep 2021.

\bibitem{isola2017image}
Phillip Isola, Jun-Yan Zhu, Tinghui Zhou, and Alexei~A Efros.
\newblock Image-to-image translation with conditional adversarial networks.
\newblock In {\em 2017 IEEE Conference on Computer Vision and Pattern
  Recognition (CVPR)}, 2017.

\bibitem{jegou2017one}
Simon J{\'e}gou, Michal Drozdzal, David Vazquez, Adriana Romero, and Yoshua
  Bengio.
\newblock The one hundred layers tiramisu: Fully convolutional densenets for
  semantic segmentation.
\newblock In {\em Proc. CVPR workshops}, pages 11--19, 2017.

\bibitem{DBLP:journals/corr/abs-1903-10663}
HeeJae Jun, ByungSoo Ko, Youngjoon Kim, Insik Kim, and Jongtack Kim.
\newblock Combination of multiple global descriptors for image retrieval.
\newblock {\em CoRR}, 2019.

\bibitem{sota}
R.~Kapoor, D.~Sharma, and T~Gulati.
\newblock State of the art content based image retrieval techniques using deep
  learning: a survey.
\newblock {\em Multimed Tools Appl}, 80:29561–29583, 2021.

\bibitem{KEIKHOSRAVI2014531}
Adib Keikhosravi, Jeremy~S. Bredfeldt, Abdul~Kader Sagar, and Kevin~W.
  Eliceiri.
\newblock Chapter 28 - second-harmonic generation imaging of cancer.
\newblock In {\em Quantitative Imaging in Cell Biology}, volume 123 of {\em
  Methods in Cell Biology}, pages 531--546. Academic Press, 2014.

\bibitem{Komura345785}
Daisuke Komura, Keisuke Fukuta, Ken Tominaga, Akihiro Kawabe, Hirotomo Koda,
  Ryohei Suzuki, Hiroki Konishi, Toshikazu Umezaki, Tatsuya Harada, and Shumpei
  Ishikawa.
\newblock Luigi: Large-scale histopathological image retrieval system using
  deep texture representations.
\newblock {\em bioRxiv}, 2018.

\bibitem{KongSRF_BMVC_2017}
Bailey Kong, James~Steven Supancic, Deva Ramanan, and Charless~C. Fowlkes.
\newblock Cross-domain forensic shoeprint matching.
\newblock In {\em British Machine Vision Conference (BMVC)}, 2017.

\bibitem{10.1155/2021/5577735}
Haopeng Lei, Simin Chen, Mingwen Wang, Xiangjian He, Wenjing Jia, and Sibo Li.
\newblock A new algorithm for sketch-based fashion image retrieval based on
  cross-domain transformation.
\newblock {\em Wireless Communications and Mobile Computing}, 2021, 2021.

\bibitem{8385104}
Yansheng Li, Yongjun Zhang, Xin Huang, and Jiayi Ma.
\newblock Learning source-invariant deep hashing convolutional neural networks
  for cross-source remote sensing image retrieval.
\newblock {\em IEEE Transactions on Geoscience and Remote Sensing},
  56(11):6521--6536, 2018.

\bibitem{tcnet}
Hangyu Lin, Yanwei Fu, Peng Lu, Shaogang Gong, Xiangyang Xue, and Yu-Gang
  Jiang.
\newblock {TC-Net for ISBIR}: Triplet classification network for instance-level
  sketch based image retrieval.
\newblock In {\em Proc. ACM Intl. Conf. on Multimedia}, page 1676–1684. ACM,
  2019.

\bibitem{9312626}
Fangcen Liu, Chenqiang Gao, Yongqing Sun, Yue Zhao, Feng Yang, Anyong Qin, and
  Deyu Meng.
\newblock Infrared and visible cross-modal image retrieval through shared
  features.
\newblock {\em IEEE Transactions on Circuits and Systems for Video Technology},
  31(11):4485--4496, 2021.

\bibitem{790410}
D.~G. {Lowe}.
\newblock Object recognition from local scale-invariant features.
\newblock In {\em Proc. Intl. Conf. on Computer Vision (ICCV)}, volume~2, pages
  1150--1157, 1999.

\bibitem{9313211}
Ashery Mbilinyi and Heiko Schuldt.
\newblock Cross-modality medical image retrieval with deep features.
\newblock In {\em Intl. Conf. on Bioinformatics and Biomedicine (BIBM)}, pages
  2632--2639, 2020.

\bibitem{cbirs}
H.~Müller, N.~Michoux, D.~Bandon, and A.~Geissbuhler.
\newblock A review of content-based image retrieval systems in medical
  applications—clinical benefits and future directions.
\newblock {\em International Journal of Medical Informatics}, 73(1):1 -- 23,
  2004.

\bibitem{lora408237}
Sebastian Ot{\'a}lora, Roger Schaer, Oscar Jimenez-del Toro, Manfredo Atzori,
  and Henning M{\"u}ller.
\newblock Deep learning based retrieval system for gigapixel histopathology
  cases and open access literature.
\newblock {\em bioRxiv}, 2018.

\bibitem{bag2}
J.~{Philbin}, O.~{Chum}, M.~{Isard}, J.~{Sivic}, and A.~{Zisserman}.
\newblock Object retrieval with large vocabularies and fast spatial matching.
\newblock In {\em 2007 IEEE Conference on Computer Vision and Pattern
  Recognition}, pages 1--8, 2007.

\bibitem{pielawski2020comir}
Nicolas Pielawski, Elisabeth Wetzer, Johan \"{O}fverstedt, Jiahao Lu, Carolina
  W\"{a}hlby, Joakim Lindblad, and Natasa Sladoje.
\newblock {CoMIR}: Contrastive multimodal image representation for
  registration.
\newblock In {\em Advances in Neural Information Processing Systems},
  volume~33, pages 18433--18444. Curran Associates, Inc., 2020.

\bibitem{10.1007/978-3-030-89128-2_28}
Lorenzo Putzu, Andrea Loddo, and Cecilia~Di Ruberto.
\newblock Invariant moments, textural and deep features for diagnostic {MR} and
  {CT} image retrieval.
\newblock In {\em Computer Analysis of Images and Patterns}, pages 287--297.
  Springer Intl. Publishing, 2021.

\bibitem{deepret}
A.~Qayyum, S.~M. Anwar, M.~Awais, and M.~Majid.
\newblock Medical image retrieval using deep convolutional neural network.
\newblock {\em Neurocomputing}, 266:8 -- 20, 2017.

\bibitem{6783793}
Atilla Sit and Daisuke Kihara.
\newblock Comparison of image patches using local moment invariants.
\newblock {\em IEEE Transactions on Image Processing}, 23(5):2369--2379, 2014.

\bibitem{bag}
J.~{Sivic} and A.~{Zisserman}.
\newblock Efficient visual search of videos cast as text retrieval.
\newblock {\em IEEE Transactions on Pattern Analysis and Machine Intelligence},
  31(4):591--606, 2009.

\bibitem{dssa}
Jifei Song, Qian Yu, Yi-Zhe Song, Tao Xiang, and Timothy~M. Hospedales.
\newblock Deep spatial-semantic attention for fine-grained sketch-based image
  retrieval.
\newblock In {\em Intl. Conf. on Computer Vision (ICCV)}, pages 5552--5561,
  2017.

\bibitem{8237837}
Ancong Wu, Wei-Shi Zheng, Hong-Xing Yu, Shaogang Gong, and Jianhuang Lai.
\newblock {RGB}-infrared cross-modality person re-identification.
\newblock In {\em Intl. Conf. on Computer Vision (ICCV)}, pages 5390--5399,
  2017.

\bibitem{8985543}
Wei Xiong, Yafei Lv, Xiaohan Zhang, and Yaqi Cui.
\newblock Learning to translate for cross-source remote sensing image
  retrieval.
\newblock {\em IEEE Transactions on Geoscience and Remote Sensing},
  58(7):4860--4874, 2020.

\bibitem{9044737}
Wei Xiong, Zhenyu Xiong, Yaqi Cui, and Yafei Lv.
\newblock A discriminative distillation network for cross-source remote sensing
  image retrieval.
\newblock {\em IEEE Journal of Selected Topics in Applied Earth Observations
  and Remote Sensing}, 13:1234--1247, 2020.

\bibitem{9222290}
Erkun Yang, Mingxia Liu, Dongren Yao, Bing Cao, Chunfeng Lian, Pew-Thian Yap,
  and Dinggang Shen.
\newblock Deep bayesian hashing with center prior for multi-modal neuroimage
  retrieval.
\newblock {\em IEEE Transactions on Medical Imaging}, 40(2):503--513, 2021.

\bibitem{10.1007/978-3-030-01216-8_19}
Jingyi Zhang, Fumin Shen, Li~Liu, Fan Zhu, Mengyang Yu, Ling Shao, Heng~Tao
  Shen, and Luc Van~Gool.
\newblock Generative domain-migration hashing for sketch-to-image retrieval.
\newblock In Vittorio Ferrari, Martial Hebert, Cristian Sminchisescu, and Yair
  Weiss, editors, {\em Computer Vision -- ECCV 2018}, pages 304--321, Cham,
  2018. Springer Intl. Publishing.

\bibitem{CrossDomainRev}
Xiaoping Zhou, Xiangyu Han, Haoran Li, Jia Wang, and Xun Liang.
\newblock Cross-domain image retrieval: methods and applications.
\newblock {\em J Multimed Info Retr}, 11:199--218, 2022.

\bibitem{CycleGAN2017}
Jun-Yan Zhu, Taesung Park, Phillip Isola, and Alexei~A Efros.
\newblock Unpaired image-to-image translation using cycle-consistent
  adversarial networks.
\newblock In {\em Intl. Conf. on Computer Vision {(ICCV)}}, 2017.

\end{thebibliography}
\end{document}